\documentclass[preprint,10pt,number,a4paper]{elsarticle}

\usepackage[ruled,linesnumbered,algo2e]{algorithm2e}
\usepackage{amsmath}
\usepackage{amssymb}
\usepackage{booktabs}
\usepackage{dsfont}
\usepackage{inputenc}
\usepackage{stmaryrd}
\usepackage{tikz}
\usepackage{url}
\usetikzlibrary{arrows,automata}

\newtheorem{definition}{Definition}[section]

\DeclareMathOperator*{\argmax}{arg\,max}

\newcommand{\naive}{na\"{i}ve}
\newcommand{\Naive}{Na\"{i}ve}

\begin{document}

\begin{frontmatter}



\title{Time and Activity Sequence Prediction of Business Process Instances}

\author[pd]{M.~Polato\corref{cor1}}
\ead{mpolato@math.unipd.it}
\author[pd]{A.~Sperduti}
\ead{sperduti@math.unipd.it}
\author[uibk]{A.~Burattin}
\ead{andrea.burattin@uibk.ac.at}
\author[tue]{M.~de Leoni\tnoteref{fn1}}
\ead{mdeleoni@m.d.leoni@tue.nl}

\cortext[cor1]{Corresponding author} 
\tnotetext[fn1]{Part of the work was conducted while affiliated with University of Padua.}
\address[pd]{University of Padua, Torre Archimede,
	via Trieste 63, 35121, Padova, Italy}
\address[uibk]{University of Innsbruck, ICT Building,
	Technikerstra\ss{}e 21a,6020, Innsbruck, Austria}
\address[tue]{Eindhoven University of Technology, Eindhoven, The Netherlands}

\begin{abstract}
The ability to know in advance the trend of running process instances, with respect to different features, such as the expected completion time, would allow business managers to timely counteract to undesired situations, in order to prevent losses. Therefore, the ability to accurately predict future features of running business process instances would be a very helpful aid when managing processes, especially under service level agreement constraints. However, making such accurate forecasts is not easy: many factors may influence the predicted features.

Many approaches have been proposed to cope with this problem but all of them assume that the underling process is stationary. However, in real cases this assumption is not always true. In this work we present new methods for predicting the remaining time of running cases. In particular we propose a method, assuming process stationarity, which outperforms the state-of-the-art and two other methods which are able to make predictions even with non-stationary processes. We also describe an approach able to predict the full sequence of activities that a running case is going to take. All these methods are extensively evaluated on two real case studies.
\end{abstract}


\end{frontmatter}


\section{Introduction}
\label{sec:intro}

An increasing number of companies are using Process Aware Information Systems (PAIS) to support their business. All these systems record execution traces, called \emph{event logs}, with information regarding the executed activities. In typical scenarios, these traces do not only contain which activities have been performed, but also additional \emph{attributes}.

The extraction of useful information out of large amount of data is the main challenge of the \emph{data mining} field. However, recently, a new topic branched off data mining: \emph{process mining} \cite{VanderAalst2011a,pmManifesto}. This latter case expects that the analyzed data are specifically referring to executions of business processes, and this assumption is made throughout the analysis phases.
Frequently, under the umbrella of the term ``process mining'', three main activities are distinguished: the discovery of models, such as control-flow, describing the actual process undergoing (namely, \emph{process discovery}); the assessment of the conformance of a given event log with respect to a given process model (namely, \emph{conformance checking}); and the extension of existing process models with additional information (namely, \emph{process enhancement}).
However, apart from these classical activities, it is also possible to exploit the event log in order to create \emph{predictive models}, i.e., models that are useful to predict future characteristics of incomplete process instances.

In general, it is useful to distinguish two types of process mining techniques, according to \emph{when} the analysis takes place: \emph{(i)} a-posteriori; and \emph{(ii)} at runtime. A-posteriori, or ``off-line'', techniques use a finite portion of historical data to extract knowledge out of it. Runtime, or ``on-line'', approaches, on the other hand, give information as long as the business process is running. It is more difficult to define approaches belonging to the latter category: the process generating the data or the frequency of events emission, may be subject to modification or drifts (which may also be seasonable). To tackle this on-line problems, it is therefore necessary to use tools able to adapt to new or unobserved scenarios.

One of the most challenging task in process mining is \emph{prediction}. Obviously, predictions are useful only if the object of such predictions have not been observed yet. For this reason, prediction \emph{per se} is, inherently, a task performed on incomplete traces, at runtime.
Moreover, the ability to predict the evolution of a running case is difficult also due to the peculiarities of each process instance and, because of the ``human factor'', which might introduce strong flexibility. 

The literature proposes plenty of works aiming at improving business processes and providing support for their execution. One of the first work that analyzes the execution duration problem is described in \cite{Reijers2006}. This particular work concentrates on cross-trained resources, but no detailed prediction algorithm is reported.
van Dongen et al., in \cite{VanDongen2008,Crooy2008}, describe a prediction model which uses all the data recorded in an event log. This approach uses non-parametric regression in order to predict the ``cycle time'' of running process instances.
The recommendation system, described by van der Aalst et al. in \cite{Schonenberg2008}, is built using historical information, and is able to predict the most likely activity that a running case is going to perform.
The TIBCO Staffware iProcess Suite \cite{Schellekens2009} is one of the first commercial tools that predicts the cycle time of running process instances. This tool simulates the complete process instance without analyzing historical data. The main building blocks of the prediction are parameters, provided by the users at ``build time'', such as the process routing or the expected duration of activities.

Recently, more sophisticated methods have been proposed, to mention some of these: transition system-based models \cite{VanderAalst2009, Folino2013, Senderovich2015278, Polato2014}, probabilistic models \cite{Lakshmanan2013, RoggeSolti20151}, queue theory based \cite{hall1990queueing,bolch2006queueing} and decision tree based approaches \cite{Leoni2014, Ghattas2014}. We will discuss these methods in more details in the next section.

In this work, we focus on predicting the remaining time of running cases. We decided to ground our approach not only on the control flow, but also on additional data that we could observe. Moreover, the system we present is also capable of dealing with unexpected scenarios and evolving conditions. With respect to our seminal work \cite{Polato2014}, here we propose an improved version of that method and we also propose two novel approaches able to overcome its limitations. In particular, we are going to define two different scenarios: in the first one we assume the process has a well defined static workflow, the same assumption made in \cite{Polato2014}, while in the latter we remove such assumption and the process is considered dynamic, e.g., the process has seasonal drift.
We will show that all the proposed approaches improve state-of-the art performances in the first scenario and one of them are also able to deal with dynamic processes. We also leverage one of these models to predict the future sequence of activity of a running case. 
We assess the prediction accuracy of our methods against the state-of-the art ones using two real-life case studies concerning a process of an Italian software company and the management of road-traffic fines by a local police office of an Italian municipality. 


\vspace{1em}
The remainder of this paper is structured as follows: Section~\ref{sec:related-work} reviews recent works concerning prediction tasks in the framework of process mining. Section~\ref{sec:background} gives some essential background on process mining and machine learning concepts used throughout the paper, while Section~\ref{sec:prediction} describes the prediction approaches. Section~\ref{sec:implementation} shows the implementation and the experimental results and finally Section~\ref{sec:conc} briefly summarizes the presented content and concludes the paper.

\section{Related Work}
\label{sec:related-work}

The first framework focused on the time perspective have been proposed by Song et al., in \cite{VanderAalst2009}. They describe a prediction approach which extracts a transition system from the event log and decorates it with time information extracted from historical cases. The transition system consists in a finite state machine built with respect to a given abstraction of the events inside the events log. The time prediction is made using the time information e.g., mean duration of specific class of cases) inside the state of the transition system corresponding to the current running instance.
The work presented in \cite{Leitner2010} considers the data perspective in order to identify SLAs (Service Level Agreement) violations. In this work, authors try to estimate the amount of unknown data, in order to improve the quality of the final prediction. The actual forecast is built using a multilayer perceptron, trained with the Backpropagation algorithm.
Two works by Folino et al. \cite{Folino2012,Folino2013} report an extended version of the technique described in \cite{VanderAalst2009}. In particular, they cluster the log traces according to the corresponding ``context features'' and then, for each cluster, they create a predictive model using the method described in \cite{VanderAalst2009}. The clustering phase is performed using predictive clustering trees. In order to propose a forecast, the approach clusters the new running instance and then uses the model belonging to the specific cluster. One of the weaknesses of these methods based on \cite{VanderAalst2009} is that they assume a static process, where the event log used for the training phase contains all the possible process behaviours. Unfortunately, in real life cases this assumption is usually not valid.
\cite{Lakshmanan2013} reports an approach which uses the \emph{Instance-specific Probabilistic Process Models} (PPM) and is able to predict the likelihood of future activities. Even though the method does not provide a time prediction, it gives to the business managers useful information regarding the progress of the process. This work also shows that the PPM built is actually Markovian.
Ghattas et al., in a recent work \cite{Ghattas2014}, exploit \emph{Generic Process Model} and decision trees, based on the process context, to provide decision criteria defined according to the actual process goals.
In \cite{Leoni2014}, de Leoni et al. propose a general framework able to find correlation between business process characteristics. In this work they manipulate the event log in order to enrich it with derived information and then generate a decision tree in order to discover correlations. In particular, one of the correlation problem suggested here is the forecast of the remaining time of running case. Being based on decision trees, numeric values need to be discretized and this lower the accuracy of the method. For this reason, they do not provide any prediction example.
Finally, in \cite{Polato2014}, Polato et al. show an approach based on \cite{VanderAalst2009} in which the additional attributes of the events are taken into account in order to refine the prediction quality. This method exploits the idea of annotating a transition system (presented in \cite{VanderAalst2009}) adding machine learning models, such as \Naive~Bayes and Support Vector Regressor. The experimental results show how the additional attributes can influence positively the prediction results.

In this paper we propose two new approaches based on Support Vector Regression and we discuss their strengths and their weaknesses comparing with the approaches presented in \cite{VanderAalst2009}. In particular we emphasize in which scenario an approach is better then the others and why.

Approaches coming from different areas can also be used to achieve similar results. For example, queue theory~\cite{hall1990queueing,bolch2006queueing} and queue mining can be seen as a very specific type of process mining, and recent works are starting to aim for similar prediction  purposes~\cite{Senderovich2015278}. In this particular case, authors decided to focus on the delay prediction problem (i.e., providing information to user waiting in lines, in order to improve customer satisfactions). The method presented here is based on the construction of an annotated transition system. Such model is then used to make delay predictions using simple averages or non-linear regression algorithms. They also show other methods based on queue mining techniques. Even if this approach concerns making predictions on the time perspective, the slant is totally different and the goal is more narrow with respect to the aim of the approach we propose in this work.
Another example of prediction-focused paper has recently been published by Rogge-Solti and Weske~\cite{RoggeSolti20151}. Authors, in this case, focused on the prediction of the remaining time, in order to avoid missing deadlines. In this case, however, a Petri net representation of the process model is needed. In the work we present, we are going to relax this assumption since we will solely rely on our log (and a transition system, built starting from it).

\section{Background}
\label{sec:background}

This section describes the basic notations and definitions necessary to understand our approach.


\subsection{Preliminaries}
\label{sub-sec:preliminaries}

A \textit{multiset} $M$ (also known as \emph{bag} or \emph{m-set}) \cite{Hickman1980} is a generalization of set in which the elements may occurs multiple times, but these are not treated as repeated elements. 
It is formally defined as a function $M : A \to \mathbb{N}^+$ such that for each $a \in A, M(a) > 0$. The set $A$ is called the \emph{root set} because an element $a$ is contained into the m-set $M$, $a \in_m M$, if $a \in A$.

The cardinality of a multiset $M : A \to \mathbb{N}^+$, denoted by $\#M$, is equal to the sum of the multiplicity of its elements, $\#M = \sum_{a \in A} M(a)$.
We call $\mathbb{B}(A) : A \to \mathbb{N}$ the set of multiset over a finite set A, i.e., $X \in \mathbb{B}(A)$ is a m-set. 

Given two multisets $M \in A \to \mathbb{N}^+, M' \in B \to \mathbb{N}^+$ the notions of intersection and disjoint union are the following (to ease the readability we assume that $X(c) = 0$ if $c \notin \textit{Dom}(X)$):
\begin{itemize}
\item \emph{Intersection}: \\
$M = X \nplus X' = \{(A \: \cap \: B) \to \mathbb{N}^+ \mid \forall \: c \in A \: \cap \: B, \: M(c) = \min(X(c), X'(c))\}$
\item \emph{Disjoint Union}: \\
$M = X \uplus X' = \{(A \: \cup \: B) \to \mathbb{N}^+ \mid \forall \: c \in A \: \cup \: B, \: M(c) = X(c) + X'(c))\}$
\end{itemize}


Let us now define the concept of sequence. Given a set $A$, a finite sequence over $A$ of length $n$ is a mapping $s \in \mathbb{S} : ([1, n] \subset \mathbb{N}) \rightarrow A$, and it is represented by a string, i.e., $s = \langle s_1, s_2, \dots, s_n\rangle$. Over a sequence $s$ we define the following functions:
\begin{itemize}
    \item \emph{selection operator} $(\cdot)$: $s(i) = s_i,\: \forall \: 1 \leq i \leq n$;
	\item $\mathit{hd}^k(s) = \langle s_1, s_2, \dots, s_{\min(k, n)} \rangle$;
	\item $\mathit{tl}^k(s) = \langle s_w, s_{w+1}, \dots, s_n \rangle$ where $w = \max(n - k + 1, 1)$;
	\item $|s| = n$;
	\item $s \uparrow A$, is the projection of $s$ onto some set $A$, e.g., \\
	$\langle a, b, b, c, d, d, d, e \rangle \uparrow \{a, d\} = \langle a, d, d, d \rangle$;
	\item $\textit{set}(s) = \{ s_i \mid s_i \in s \}$, e.g., $\textit{set}(\langle a, b, b, c, d, d, d, e \rangle) = \{a, b, c, d, e\}$.
\end{itemize}

\subsection{Event Logs}
\label{sub-sec:event-logs}

Process mining techniques can extract information from event logs (Table~\ref{tb:log}). Usually, these techniques assume that event logs are well structured, in particular, they assume that each event of a business process is recorded and it refers to an actual activity of a particular case. Even other additional information may be required by a process mining algorithm, such as the originator of an activity or the timestamp. Nowadays, many companies are using softwares which keep tracks of the business process execution in the form of event logs (e.g., transaction logs, databases, spreadsheets, etc). 

In this section we define some useful process mining concepts which we will use throughout the paper.
First of all, we give the basic definition of \emph{event}, \emph{trace} and \emph{event log}.

\begin{definition}[Event]
    An \emph{event} is a tuple $e = (a, c, t, d_1, \dots, d_m)$, where $a \in \mathcal{A}$ is the process activity associated to the event, $c \in \mathcal{C}$ is the \emph{case id}, $t \in \mathbb{N}$ is the event timestamp (seconds since 1/1/1970\footnote{We assume this representation according to the Unix epoch time.}) and $d_1, \dots, d_m$ is a list of additional attributes, where $\forall \: 1 \leq i \leq m, \: d_i \in \mathcal{D}_i$.\\
    We call $\mathcal{E} = \mathcal{A} \times \mathcal{C} \times \mathcal{T} \times \mathcal{D}_1 \times \dots \times \mathcal{D}_m$ the \emph{event universe}.
\end{definition}

Over an event $e$ we define the following \emph{projection} functions: $\pi_\mathcal{A}(e) = a$, $\pi_\mathcal{C}(e) = c$, $\pi_\mathcal{T}(e) = t$ and \mbox{$\pi_{\mathcal{D}_i}(e) = d_i, \forall \: 1 \leq i \leq m$}. If $e$ does not contain the attribute value $d_i$ for some $i \in [1, m] \subset \mathbb{N}$, $\pi_{\mathcal{D}_i}(e) = \perp$. In the remainder of this paper we will call the number of additional attribute $m$ as $|\mathcal{D}|$.

\begin{definition}[Trace, Partial Trace]
    A \emph{trace} is a finite sequence of events $\sigma_c = \langle e_1, e_2, \dots, e_{|\sigma_c|} \rangle \in \mathcal{E}^*$ such that $\forall \: 1 \leq i \leq |\sigma_c|, \pi_\mathcal{C}(e_i) = c \wedge \forall \: 1 \leq j \leq |\sigma_c|$, \mbox{$\pi_\mathcal{T}(\sigma_c(j)) \leq \pi_\mathcal{T}(\sigma_c(j + 1))$}. We define a \emph{partial trace} of length $k$ as $\sigma^k_c = hd^k(\sigma_c)$, for some $k \in [1, |\sigma_c|] \subset \mathbb{N}$.
We call $\Sigma$ the set of all possible (partial) traces.
\end{definition}

While a trace corresponds to a complete process instance, i.e., an instance which is both started and completed; a partial trace represents a process instance which is still in execution and hence it has not completed yet. Over a trace $\sigma_c = \langle e_1, e_2, \dots, e_{|\sigma_c|} \rangle$ we define the following projection functions: $\Pi_\mathcal{A}(\sigma_c) = \langle \pi_\mathcal{A}(e_1), \pi_\mathcal{A}(e_2), \dots, \pi_\mathcal{A}(e_{|\sigma_c|}) \rangle$, $\Pi_\mathcal{T}(\sigma_c) = \langle \pi_\mathcal{T}(e_1), \pi_\mathcal{T}(e_2), \dots, \pi_\mathcal{T}(e_{|\sigma_c|}) \rangle$ and $\Pi_{\mathcal{D}_i}(\sigma_c) = \langle \pi_{\mathcal{D}_i}(e_1), \pi_{\mathcal{D}_i}(e_2), \dots, \pi_{\mathcal{D}_i}(e_{|\sigma_c|}) \rangle$ for all $1 \leq i \leq |\mathcal{D}|$.

Let us now define the concept of event log as in \cite{VanderAalst2009}.

\begin{definition}[Event log]
An \emph{event log} $L$ is a set of traces, $L = \{\sigma_c \mid c \in \mathcal{C}\}$ such that each event appears at most once in the entire log, i.e., $\forall \sigma_1, \sigma_2 \in L, \sigma_1 \neq \sigma_2 : \textit{set}(\sigma_1) \cap \textit{set}(\sigma_1) = \emptyset$.
\end{definition}

\begin{table}
\centering
\begin{small}
	\begin{tabular}{c|c|c|c|c|c}
		\toprule
		\textbf{Case Id}&\textbf{Timestamp}&\textbf{Resource}&\textbf{Activity}&\textbf{Category}&\textbf{Amount}\\
		\midrule
		65923  & 20-02-2002:11.11 & Jack & A & - & 1000  \\
		65923  & 20-02-2002:13.31 & Jack & B & Gold & 1000 \\
		65923  & 21-02-2002:08.40 & John & C & Gold & 900 \\
		65923  & 22-02-2002:15.51 & Joe & F & Gold & 900 \\
		\midrule
		65924  & 19-02-2002:09.10 & Jack & A & - & 200 \\
		65924  & 19-02-2002:13.22 & John & B & Standard & 200 \\
		65924  & 20-02-2002:17.17 & John & D & Standard & 200 \\
		65924  & 21-02-2002:10.38 & Joe & F & Standard  & 200 \\
		\midrule
		65925  & 25-02-2002:10.50 & Jack & A & - & 850 \\
		65925  & 25-02-2002:13.01 & John & B & Gold & 850 \\
		65925  & 25-02-2002:16.42 & Joe & E & Gold & 500 \\
		65925  & 26-02-2002:09.30 & Joe & F & Gold  & 500 \\
		\bottomrule
	\end{tabular}
\end{small}
\caption{Example of an event log fragment with events sorted using their Timestamp and grouped by Case Id.}
\label{tb:log}
\end{table}

\subsection{Transition System}
\label{sub-sec:transition-system}

With the definitions given in the previous subsections we can now characterize the concept of transition system and how to construct it starting from an event log.

A transition system is one of the simplest process modeling notations, it consists of \emph{states} and \emph{transitions} where each transition connect two states (not necessarily different). A transition system is also referred as a Finite-State Machine (FSM). From a mathematical point of view, it can be seen as a directed graph in which every possible path from the initial state to the accepting ones represents a possible behavior of the underlying process.

Formally, it is defined as follows:
\begin{definition}[Transition System (TS)]
A \emph{transition system} is a triplet $\textit{TS} = (S, A, T)$, where $S$ is the set of \emph{states}, $A \subseteq \mathcal{A}$	is the set of activities and $T \subseteq S \times A \times S$ is the set of \emph{transitions}. $S^{\textit{start}} \subseteq S$ is the set of \emph{initial states}, and $S^{\textit{end}}  \subseteq S$ is the set of \emph{final (accepting) states}.
\end{definition}

A \emph{walk}, in a transition system, is a sequence of transitions $\langle t_1, t_2, \dots, t_n \rangle$ such that \mbox{$t_1 = (s_1 \in S^{\textit{start}}, e, s'_1), \: t_n = (s_n, e, s'_n\in S^{\textit{end}}) $} and $\forall  \: 1 < h < n, \: t_h = (s_h, e, s_{h+1})$.
Given a state $s \in S$, it is possible to define the set of reachable states from $s$ as: $s\bullet = \{s' \in S\: |\: \exists t \in T,  \exists e \in E\: \text{s.t.}\: t = (s, e, s') \}$. 

According to van der Aalst et al. \cite{VanderAalst2009}, to construct a transition system which maps each partial trace in the log to a state, we need the so called \emph{state} and \emph{event representation functions}.

\begin{definition}[State representation function]
	Let $\mathcal{R}_s$ be the set of possible state representations, a state representation function $f^{\textit{state}} \in \Sigma \rightarrow \mathcal{R}_s$ is a function that, given a (partial) trace $\sigma$ returns some representation of it (e.g., sequences, sets, multiset over some event properties).
\end{definition}

\begin{definition}[Event representation function]
	Let $\mathcal{R}_e$ be the set of possible event representations, an event representation function $f^{\textit{event}} \in \mathcal{E} \rightarrow \mathcal{R}_e$ is a function that, given an event $e$ produces some representation of it (e.g., projection functions over $e \in \mathcal{E}$: $\pi_\mathcal{A}(e), \pi_\mathcal{T}(e)$).
\end{definition}

Choosing the right functions $f^{\textit{state}}$ and $f^{\textit{event}}$, also referred to as \emph{abstractions}, is not a trivial task \cite{VanderAalst2008, VanderAalst2009}. A conservative choice (e.g., no abstraction: \mbox{$f^{\textit{state}}(\sigma_c) = \sigma_c$}, \mbox{$f^{\textit{event}}(e) = e$}) can lead to a transition system which does \emph{overfit} the log $L$, because the state space becomes too large and specific. An aggressive choice (e.g., \mbox{$f^{\textit{state}}(\sigma_c) = \{\sigma_c(|\sigma_c|)\}$}), instead, can lead to a transition system that \emph{overgeneralizes} the log $L$, allowing too much behaviors. In this latter case the transition system is \emph{underfitting} $L$. Some possible good choices for $f^{\textit{state}}$ and $f^{\textit{event}}$ are described and discussed in \cite{VanderAalst2008} and \cite{VanderAalst2009}. A common event abstraction is $f^{\textit{event}}(e) = \pi_\mathcal{A}(e)$, which maps an event onto the name of the activity, while commons state abstractions are: the \emph{set abstraction}, i.e., {$f^{\textit{state}}(\sigma_c) = \{\pi_\mathcal{A}(e) \mid e \in \sigma_c\}$}, the \emph{multiset abstraction}, i.e., {$f^{\textit{state}}(\sigma_c) = \{(a, m) \mid a = \pi_\mathcal{A}(e) \wedge m = |\Pi_\mathcal{A}(\sigma_c) \uparrow \{a\}| \}$} and the \emph{list abstraction}, i.e., {$f^{\textit{state}}(\sigma_c) = \langle \pi_\mathcal{A}(\sigma_c(1)), \dots, \pi_\mathcal{A}(\sigma_c(|\sigma_c|)) \rangle$}.

Using these two functions $f^\textit{event}$ and $f^\textit{state}$, it is possible to define a (labeled) transition system where states correspond to prefixes of traces in the log mapped to some representations using $f^\textit{state}$, and transitions correspond to representation of events through $f^\textit{event}$. 

\begin{definition}[Labeled Transition System (LTS)]
	Given a state representation function $f^{\textit{state}}$, an event representation function $f^{\textit{event}}$ and an event log $L$, we define a Transition System as $\textit{LTS} = (S, E, T)$, where:
	\begin{itemize}
		\item $S = \{f^{\textit{state}}(\mathit{hd}^k(\sigma)) \mid \sigma \in L \wedge 0 \leq k \leq |\sigma|\}$ is the state space;
		\item $E = \{f^{\textit{event}}(\sigma(k)) \mid \sigma \in L \wedge 1 \leq k \leq |\sigma|\}$ is the set of event labels;
		\item $T (\subseteq S \times E \times S) = \{f^{\textit{state}}(hd^k(\sigma)), \: f^{\textit{event}}(\sigma(k+1)), \: f^{\textit{state}}(hd^{k+1}(\sigma))) \: |$ \mbox{$\sigma \in L \wedge 0 \leq k < |\sigma| \}$} is the transition relation.
	\end{itemize}
	$S^{\textit{start}} = \{f^{\textit{state}}(\langle\rangle)\}$ is the set with the unique initial state, and $S^{\textit{end}} = \{f^{\textit{state}}(\sigma) \mid \sigma \in L\}$ is the set of final (accepting) states.
\end{definition}

We say that a trace is \emph{compliant} with the transition system if it corresponds to a walk from $s_i \in S^{\textit{start}}$ to $s_e \in S^{\textit{end}}$. We also call a trace $\sigma$ \textit{non-fitting} with respect to a transition system if $f^\textit{state}(\sigma) \notin S$.

\begin{figure}[h]
\centering
\begin{tikzpicture}[>=stealth',shorten >=1pt,auto,node distance=2.5cm]
    \node[state] (o)                {${s_0}_{\{\}}$};
    \node[state] (A) [right of = o] {${s_1}_{\{A\}}$};
    \node[state] (B) [right of = A] {${s_2}_{\{B\}}$};
    \node[state] (D) [right of = B] {${s_3}_{\{D\}}$};
    \node[state] (C) [above of = D] {${s_4}_{\{C\}}$};
    \node[state] (E) [below of = D] {${s_5}_{\{E\}}$};
    \node[state,accepting] (F) [right of = D] {${s_6}_{\{F\}}$};
    \path[->](o) edge node [align=center]{$A$} (A) 
             (A) edge node [align=center]{$B$} (B)
             (B) edge node [align=center]{$C$} (C)
             (B) edge node [align=center]{$D$} (D)
             (B) edge [below left] node [align=center]{$E$} (E)
             (C) edge node [align=center]{$F$} (F)
             (D) edge node [align=center]{$F$} (F)
             (E) edge [below right] node [align=center]{$F$} (F);
\end{tikzpicture}
\caption{Example of a transition system extracted from a log containing three trace types $\langle A, B, C, F\rangle$, $\langle A, B, D, F\rangle$ and $\langle A, B, E, F\rangle$, with $f^\textit{event}(e) = \pi_\mathcal{A}(e)$ and $f^\textit{state}(\sigma) = \{f^\textit{event}(\sigma(|\sigma|))\}$. The state $s_0$ is the initial state, while $s_6$ is the accepting (i.e., final) state. The notation ${s_1}_{\{A\}}$ means that the state $s_1$ has a state representation equals to $\{A\}$. Each transition is labeled by the corresponding event representation value.\label{fig:ts}}
\end{figure}
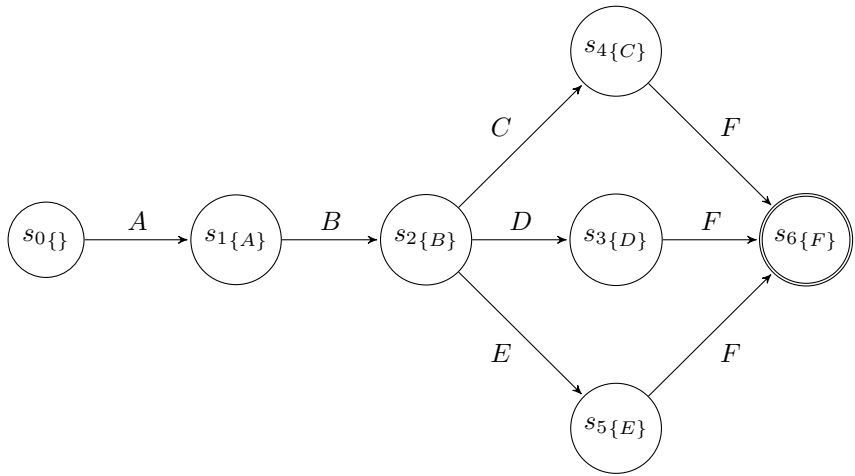

A straightforward method for constructing a transition system, given a log $L$, is the following: for each trace $\sigma \in L$, and for each $ 1 \leq k \leq |\sigma|$, we create, if does not exist yet, a new state $f^\textit{state}(\textit{hd}^k(\sigma))$. Then, through a second iteration over $k$, $ 1 \leq k < |\sigma|$, we create, if does not exist yet, a new transition $f^\textit{event}(\sigma(k + 1)) \mid f^\textit{state}(\textit{hd}^k(\sigma)) \xrightarrow{f^\textit{event}(\sigma(k+1))} f^\textit{state}(\textit{hd}^{k+1}(\sigma))$. Algorithm~\ref{alg:ts-construction} shows a pseudocode of the method just described, while Figure~\ref{fig:ts} depicts an example of a transition system extracted from the log fragment reported in Table~\ref{tb:log}.

\begin{algorithm2e}
    \caption{Construction of a Transition System\label{alg:ts-construction}}
    \SetKwComment{tcp}{$\triangleright\ $}
	\DontPrintSemicolon
	\KwIn{$L$: event log; $f^{\textit{state}}$: state representation function;  $f^{\textit{event}}$: event representation function}
	\KwOut{$\textit{TS}$: transition system}
	\BlankLine
	\DontPrintSemicolon
	$E, T \gets \emptyset$ \;
	$S \gets \{f^\textit{state}(\langle\rangle)\}$ \tcp*[r]{initialize with the start state}
	\BlankLine
    \ForEach{$\sigma \in L$} {
		\For{$k  \gets 1\ \mbox{\bf to}\ |\sigma|$}{
    		\If{ $s = f^{\textit{state}}(hd^k(\sigma)) \notin S$} {
                $S \gets S \cup \{s\}$ \tcp*[r]{add a new state if necessary}
    		}
    	}
    }
	\BlankLine
	\ForEach{$\sigma \in L$} {
		\For {$k \gets 0\ \mbox{\bf to}\ |\sigma| - 1$} {
		     $s \gets f^{\textit{state}}(\mathit{hd}^k(\sigma))$ \;
		     $e \gets f^{\textit{event}}(\sigma(k+1))$ \;
		     $s' \gets f^{\textit{state}}(\mathit{hd}^{k+1}(\sigma)))$ \;
		\BlankLine		     
		     \If { $e \notin E$} {
                $E \gets E \cup \{e\}$
    		}
    	\BlankLine
    		\If { $t = (s,\:e,\:s') \notin T$} {
                $T \gets T \cup \{t\}$ \tcp*[r]{add a new transition if necessary}
    		}
    	}
    }
    $\textit{TS} \gets (S, E, T)$ \;
    \KwRet{$\textit{TS}$}
\end{algorithm2e}

\section{Machine Learning Background}
\label{sec:machine_learning}

This section provides the fundamental machine learning concepts required throughout the paper.

\subsection{\Naive~Bayes Classifier}
\label{sub-sec:bayes}

\emph{Classification} is a machine learning task that consists in predicting category membership of data instances. More formally, given a ``concept'' $\mathcal{F} : X \rightarrow Y$ which maps elements of the domain $X$ into a range $Y = \{y_1, y_2, \dots, y_m\}$ (i.e., the possible categorizations), the classification task consists in learning a function $\tilde{\mathcal{F}}$ which constitutes a good approximation of $\mathcal{F}$.

\Naive~Bayes (NB) \cite{Mitchell} is a probabilistic classifier which is based on the application of Bayes' theorem. This classifier belongs to the family of the so called \emph{supervised algorithms}. These algorithms need a set of pre-classified instances in order to learn how to classify new, unseen, instances.

Let $\vec{x} = (x_1, x_2, \dots, x_n) \in X$ be an $n$-dimensional vector. From a probabilistic point of view, the probability that $\vec{x}$ belongs to a category $y_i \in Y$ is given by the Bayes' theorem:
$
	P(y_i \: | \: \vec{x}) = P(y_i) P(\vec{x} \: | \: y_i) / P(\vec{x})
$,
where $P(y_i)$ is the a-priori probability of $y_i$ and $P(\vec{x})$ is the a-priori probability of $\vec{x}$. The estimation of the conditional probability $P(\vec{x} \: | \: y_i)$ can be very hard to compute, because of the number of possible $\vec{x}$ may be very large. In order to simplify this problem, we can assume that the components $x_i$ of the vector $\vec{x}$ (viewed as random variables) are conditionally independent each other given the target information $y_i$. With this assumption, $P(\vec{x} \: | \: y_i)$ can be easily computed by
this product $\prod_{k = 1}^n P(x_k \: | \: y_i)$.
However, the independence assumption is quite strong and in many cases it does not hold. This is why this method is usually named \emph{\naive}.

To get the classification of the vector $\vec{x}$, we have to find out the \textit{maximum a posteriori (MAP)} class $y_{\textit{MAP}}$ :
\begin{align}
     y_{\textit{MAP}} &= \argmax_{y \in Y} \: P(y \: | \: x_1, x_2, \dots, x_n)\\
             &= \argmax_{y \in Y} P(y) \prod_{k = 1}^n P(x_k \: | \: y).
\end{align}
It is worth to notice that in (2), if one (or more) probability $P(x_k \: | \: y)$ is zero the whole product is equal to zero. This undesirable situation can be avoid applying the Laplacian (or additive) smoothing \cite{Mitchell} to the conditional probabilities $P(x_k \: | \: y)$. Sometimes, it is necessary to get not only the MAP class, but the probability distribution of the whole category. This can be compute by $P(y) \prod_{k = 1}^n P(x_k \: | \: y) / P(x_1, \dots, x_n)$ for all $y \in Y$.

The training phase of this method consists in the collection of the statistics, from the training set, necessary to calculate $y_{\textit{MAP}}$. The classification is simply the application of $y_{\textit{MAP}}$ (or the class distribution) over the input vectors.

\subsection{Support Vector Regression}
\label{sub-sec:svr}

Regression analysis is a statistical process for estimating the relationships among variables. This approach is widely used for prediction and forecasting. One of the most recently proposed approaches is the Support Vector Regression (SVR) \cite{Basak2007,Drucker1997,Smola1998}, which is, as the name suggests, based on Support Vector Machines (SVM). 

Let $\textit{Tr} = \{(\vec{x}_1, y_1), (\vec{x}_2, y_2), \dots, (\vec{x}_l, y_l)\} \in X \times \mathbb{R}$ be the training data, where $X \equiv \mathbb{R}^n$, for some $n \in \mathbb{N}^+$, denotes the space of the input vectors.
In $\epsilon$-SVR \cite{Drucker1997}, the goal is to find a function $f(\vec{x})$ that deviates from the target $y_i$ by at most $\epsilon$, for all the training instances. In addition to that, the function $f$ has to be as ``flat'' as possible. Let's start considering the linear case, in which $f$ has the following form:
\begin{align}\label{eq:svmplan}
    f(\vec{x}) = \langle \vec{w}, \vec{x} \rangle + b \;\; \mbox{with} \;\vec{w} \in X, b \in \mathbb{R}
\end{align}
where $\langle \vec{w}, \vec{x} \rangle$ is the dot product between $\vec{w}$ and $\vec{x}$ in $X$.
In Eq.~\ref{eq:svmplan}, \emph{flatness} means that the norm of $\vec{w}$ has to be as small as possible. A possible way to cope with this problem is to minimize the Euclidean norm, i.e., $\|\vec{w}\|^2$. Formally, this can be written as a quadratic constrained optimization problem:
$$
	\textit{minimize} \quad \frac{1}{2} \|\vec{w}\|^2 + C \sum_{i=1}^l (\xi_i+ \xi^*_i)
$$
$$
	\text{subject to} \quad \left\{
		\begin{array}{l}
			y_i - \langle \vec{w}, \vec{x}_i \rangle - b \leq \epsilon + \xi_i\\
			\langle \vec{w}, \vec{x}_i \rangle + b - y_i \leq \epsilon + \xi_i^*\\
    		\xi_i, \xi^*_i \geq 0
		\end{array}
	\right.
$$
where $\xi_i, \xi^*_i$ are \emph{slack variables} which allow the violation of some constraints and the constant $C > 0$ represents a trade-off between the amount of allowed deviations, greater than $\epsilon$, and the flatness of $f$. 
The dual formulation of this convex optimization problem \cite{Drucker1997,Basak2007,Smola1998} provides the key for extending SVR to nonlinear functions:
\begin{equation} \label{eq:lagrangian}
\begin{split}
		L = &\frac{1}{2} \|\vec{w}\| + C \sum_{i=1}^l (\xi_i+\xi_i^*) - \sum_{i=1}^l \alpha_i (\epsilon + \xi_i - y_i + \langle \vec{w}, \vec{x}_i \rangle + b) +\\ 
		&-\sum_{i=1}^l \alpha_i^* (\epsilon + \xi_i - y_i + \langle \vec{w}, \vec{x}_i \rangle + b) - \sum_{i=1}^l (\eta_i \xi_i + \eta_i^* \xi_i^*)
\end{split}
\end{equation}

where the dual variables have to satisfy positivity constraints (\mbox{$\alpha_i, \alpha_i^*, \eta_i, \eta_i^* \geq 0$}).
The optimal solution of this problem is guaranteed to be in the saddle point, hence the solution can be found setting the partial derivatives of L, with respect to the primal variables $\vec{w}, b, \xi_i, \xi_i^*$, to zero, and then substituting back on Eq.~\ref{eq:lagrangian} obtaining:
\begin{equation} \label{eq:last-svr}
\begin{split}
	\textit{maximize} \quad
    		\sum_{i, j = 1}^l &(\alpha_i - \alpha_i^*)(\alpha_j-\alpha_j^*)\langle \vec{x}_i, \vec{x}_j\rangle
    		-\epsilon \sum_{i=1}^l (\alpha_i+\alpha_i^*)+\sum_{i=1}^l y_i(\alpha_i-\alpha_i^*)\\
	&\text{subject to} \quad \left\{
		\begin{array}{l}
			\sum_{i=1}^l (\alpha_i+\alpha_i^*) = 0\\
			\alpha_i, \alpha^*_i \in [0, C].
		\end{array}
		\right.
\end{split}
\end{equation} 

The resulting hyperplane equation is $\vec{w} = \sum_{i=1}^l (\alpha_i^* - \alpha_i) \vec{x}_i$, and consequently we can rewrite Eq.~\ref{eq:svmplan} as $f(\vec{x}) = \sum_{i=1}^l (\alpha_i - \alpha_i^*)\langle \vec{x}_i, \vec{x} \rangle + b$, which is called \emph{support vector expansion}, because $\vec{w}$ is described as a linear combination of the training data $\vec{x}_i$. Exploiting the Karush-Kuhn-Tucker (KKT) conditions \cite{Smola1998}, which states that at the optimum the product between constraints and dual variables is zero, we can compute $b$ and we can also notice that only for $|f(\vec{x}_i - y_i)| = \epsilon$ the coefficients $\alpha_i, \alpha_i^*$ are non-zero. This particular set of training examples are called \emph{support vectors}.

The nonlinearity of this model can be achieved mapping ($\Phi : X \rightarrow \mathcal{S}$) the input vector $\vec{x}$ to a highly dimensional space $\mathcal{S}$ (i.e. \emph{feature space}) and then applying the standard SVR method. Cover's Theorem \cite{Cover1965} proves that given a set of training data that is not linearly separable in the input space, it is possible to transform the data into a training set that is linearly separable, with high probability, by projecting it into a higher-dimensional space (\emph{feature space}) via some non-linear transformation. 

Unfortunately a direct approach is, in most cases, infeasible from a computational point of view.
Anyhow, it is worthwhile to note that the dual form depends only on the dot product between the input vectors. Hence, to get the nonlinearity, it is sufficient to know the dot product of the input vectors in the feature space $k(\vec{x}, \vec{x}') = \langle \Phi(\vec{x}), \Phi(\vec{x}') \rangle$. Then, it is easy to rewrite the dual form with $k(\vec{x}, \vec{x}')$ instead of $\langle \vec{x}_i, \vec{x}_j\rangle$. Exploiting the Mercer's Theorem \cite{Smola1998} it is possible to characterize these type of functions $k$, called \emph{kernel functions}.
Using $\Phi$ and the kernel function $k$ we can rewrite the optimization problem and the support vector expansion substituting the products $\langle \cdot, \cdot \rangle$ with $k(\cdot, \cdot)$ and $\vec{x}_i$ with $\Phi(\vec{x}_i)$. Solving this new optimization problem, which coincides with the algorithm's training phase, means finding the flattest function in the feature space. Once defined a suitable kernel function $k$, the standard SVR function which solve the approximation problem has this form:
\begin{equation} \label{eq:svr-model}
	f^*(\vec{x}) = \sum_{i=1}^l (\alpha_i^*  -\alpha_i) k(\vec{x}_i, \vec{x}) + b
\end{equation}
where $\alpha_i$ and $\alpha_i^*$ are Lagrange multipliers obtained by solving Eq.~\ref{eq:last-svr}.

\section{Remaining Time Prediction}
\label{sec:prediction}

In this section we are going to show a spectrum of different approaches able to predict the remaining time of running business process instances. In particular, we will emphasize the pros and cons of each approach and which are the situations in which an approach should be preferred among the others. \\

The problem of predicting the remaining time of running process instances can be summarized as follows: given an event log, containing historical traces about the execution of a business process, we want to predict, for an ongoing process instance, how much time remains until its completion. 

All the approaches described in this work are based on the idea of making prediction using a model constructed (i.e., learned) using the information collected so far (i.e., the event log). The obtained model takes the partial trace, which represents the running process instance, as input and returns the remaining time forecast.

The remaining time of a case consists of the time that will be spent from now up to the end of the execution of the last activity of the process. This amount of time, given a complete trace $\sigma_c$, is easily computable for each event $e_i \in \sigma_c$. We define the function $\textit{rem} : \Sigma \times \mathbb{N} \to \mathbb{N}$ as follow:
$$
	\textit{rem}(\sigma_c, i) = \pi_\mathcal{T}(\sigma_c(|\sigma_c|)) - \pi_\mathcal{T}(\sigma_c(i))
$$
where $i$ is an event index. If $\sigma_c = \langle\rangle$ or $i \notin \{1, 2, \dots, |\sigma_c|\}$ then $\textit{rem}(\sigma_c, i) = 0$. This function calculates the time difference between the last event in the trace and the $i$-th event.

In the reminder of this section we are going to present our new time prediction approaches based on the application of machine learning models. Moreover, we discuss how to exploit one of the proposed model to predict also the future sequence of activities.

\subsection{\textbf{Approach 1}: Simple Regression}
\label{sub-sec:simple-regression}
This prediction task has all the characteristics to be faced as a regression problem. (Partial) Traces in the event log are the input vectors and the remaining time at each event are the target values.
In this section we are going to present a direct application of $\epsilon$-SVR algorithm. 
Despite the simplicity of this method, there are some aspects which need particular attention. First of all we describe how to switch from the event log to a representation suitable for the $\epsilon$-SVR algorithm.

\subsubsection{Features Representation}
\label{sub-sub-sec:feature-rep}
 
In our setting, the input consists of traces and, in particular, the attributes of the corresponding events. Whereas SVR takes as input vectors $\vec{x} \in \mathbb{R}^l$, for some $l \in \mathbb{N}^+$, we have to convert sequences of events into some representation in $\mathbb{R}^l$. 

Let us consider a (partial) trace $\sigma_c = \langle e_1, e_2, \dots, e_n\rangle$ of length $n \in \mathbb{N}^+$, for each event $e_i = (a^i, c, t^i, d_1^i, \dots, d_m^i) \in \sigma_c^k$, the attributes $d_1^i, \dots, d_m^i$ may have different values, because they can change as the process instance evolves. We consider as additional attributes values (i.e., $d_1^i, \dots, d_m^i$) the last values chronologically observed. Formally, we define the function $\textit{last} : \Sigma \times \mathbb{N} \to \mathcal{D} \cup \{\perp\}$ as:
$$
	\textit{last}(\sigma_c, i) = \left\{\pi_{\mathcal{D}_i}(e_j) \mid j = \argmax_{1\leq h \leq |\sigma_c|} \:\pi_{\mathcal{D}_i}(e_h) \neq \perp \right\}.
$$
where $i$ is the index of the attribute. If there is no index $j$ such that $\pi_{\mathcal{D}_i}(\sigma_c(j)) \neq \perp$ then $\mathit{last}(\sigma_c, i) = \perp$. What we need to do next is to transform the domains $\mathcal{D}_i$ into a numerical representation that can be given as input to the SVR. In order to do that, we use the \textit{one-hot} encoding. This encoding converts the nominal data value $d_i$ into a binary vector $v \in \{0, 1\}^{|\mathcal{D}_i|}$, with $|\mathcal{D}_i|$ components and with all values set to 0 except for the component referring to $d_i$, which is set to 1. 
All the other attributes values $d_i$ such that $\mathcal{D}_i \subseteq \mathbb{R}$ are simply put in a single component vector, e.g., let $\mathcal{D}_i \equiv \mathbb{N} \subset \mathbb{R}$ and $\pi_\mathcal{D}(e) = 17$, the output vector is $\vec{u} = [17] \in \mathbb{R}$. In the remainder of this paper we will call $\mathds{1} : A \to \mathbb{R}^j$, for some $j \in \mathbb{N}^+$, the function which maps an attribute value to its one-hot encoded vector.\\

After the conversion just described, all the vectors are concatenated (keeping the same fixed order) together, e.g., recalling $v \in \mathbb{R}^4$ and $u \in \mathbb{R}$ from the previous examples, their concatenation is equal to $\vec{z} = \vec{v} \: || \: \vec{u} = [0, 1, 0, 0]||[17] = [0, 1, 0, 0, 17] \in \mathbb{R}^5$. Note that if $\pi_{\mathcal{D}_i}(e) = \perp$ and $\mathcal{D}_i$ is nominal, then $\perp$ is projected to a vector ($\in \{0, 1\}^{|\mathcal{D}_i|}$) with all components set to zero. Otherwise, if $\mathcal{D}_i \subseteq \mathbb{R}$ the value $\perp$ is simply interpreted as a zero. Eventually, the concatenation of all vectors constitutes an input vector $\vec{x} \in \mathbb{R}^l$ for the $\epsilon$-SVR. 

We summarize all of these steps with the function $\gamma^* : \Sigma \to \mathbb{R}^l$, i.e., $\forall i, \gamma^*(\sigma^i_c) = \vec{x}_i$, such that $\gamma^*(\sigma) = \bigparallel_j \mathds{1}(\textit{last}(\sigma, j))$.

With respect to the target value, it is calculated using the function $\textit{rem}$, e.g., $y = \textit{rem}(\sigma_c, i)$ for some $1 \leq i \leq |\sigma_c|$. Hence, starting from a trace $\sigma_c = \langle e_1, e_2, \dots, e_n\rangle$, the corresponding set of $n$ training examples 
$(\vec{x}, y)$ will be $(\gamma^*(\sigma^i_c), \textit{rem}(\sigma_c, i)), \forall i \in \{1, \dots, n\}$.


\subsubsection{Training}
\label{sub-sub-sec:simple-svr-training}
As discussed in Section~\ref{sub-sec:svr}, a training data set for a $\epsilon$-SVR algorithm is defined as $\textit{Tr} = \{(\vec{x}_1, y_1), (\vec{x}_2, y_2), \dots, (\vec{x}_l, y_l)\} \in \mathbb{R}^n \times \mathbb{R}$, for some $n \in \mathbb{N}^+$. In order to map the data contained in an event log $L$, we exploit the transformations described in the previous section. In particular, the training set is created by the Algorithm~\ref{alg:training-construction}.

\begin{algorithm2e}
    \caption{Training set construction\label{alg:training-construction}}
    \SetKwComment{tcp}{$\triangleright\ $}
	\DontPrintSemicolon
	\KwIn{$L$: event log}
	\KwOut{$\textit{Tr}$: training set}
	\DontPrintSemicolon
	\BlankLine
	$\textit{Tr} \gets \emptyset$ \;
    \ForEach{$\sigma \in L$} {
		\For{$k  \gets 1\ \mbox{\bf to}\ |\sigma|$} { 
			$\vec{x} \gets \mathds{1}(\pi_\mathcal{A}(\sigma(k)))$ \label{alg-line:from} \;
    		\For(\tcp*[f]{for each attribute in $L$}){$i \gets 1\ \mbox{\bf to}\ |\mathcal{D}|$} {
    			$v = \textit{last}(\sigma^k, i)$\;
    			$\vec{x} \gets \vec{x} \: || \: \mathds{1}(v)$\; 	
    		} \label{alg-line:to}
    		$y = \textit{rem}(\sigma, k)$\;
    		$\textit{Tr} \gets \textit{Tr} \cup (\vec{x}, y)$\;
    	}
    }
    \KwRet{TS}
\end{algorithm2e}

The value returned by the function $\textit{rem}$ depends on the time granularity (e.g., hours, minutes, seconds). It is important to keep the same granularity for all the instances.
Once constructed the training set $\textit{Tr}$, the training phase consists in solving the optimization problem (Eq.~\ref{eq:last-svr}) with input $\textit{Tr}$.

\subsubsection{Prediction}
\label{sub-sub-sec:simple-svr-prediction}

After the training phase, the $\epsilon$-SVR model is created (i.e., function $f^*$, Eq.~\ref{eq:svr-model}) and can be directly used to predict the remaining time of partial traces. First of all the trace is converted to a vector $\vec{x}$ suitable for the SVR, applying the same approach illustrated in Algorithm~\ref{alg:training-construction}, in particular from line~\ref{alg-line:from} to line~\ref{alg-line:to}. Then this vector $\vec{x}$ is given as input to $f^*$ which produces the time prediction. This prediction value has to be interpreted with the same granularity used in the training instances creation. Algorithm~\ref{alg:svr-prediction} shows the prediction algorithm.

\begin{algorithm2e}
    \caption{Prediction\label{alg:svr-prediction}}
    \SetKwComment{tcp}{$\triangleright\ $}
	\DontPrintSemicolon
	\KwIn{$\sigma_p$: (partial) trace, $f^*$: $\epsilon$-SVR model}
	\KwOut{$\textit{P}$: time prediction}
	\DontPrintSemicolon
	\BlankLine
    $\vec{x} \gets \mathds{1}(\pi_\mathcal{A}(\sigma_p(|\sigma_p|)))$\; \label{alg-line:first}
    \For{$i \gets 1\ \mbox{\bf to}\ |\mathcal{D}|$} {
    	$v = \textit{last}(\sigma_p^k, i)$\;
    	$\vec{x} \gets \vec{x} \: || \: \mathds{1}(v)$\; 	
    }
	\BlankLine
	$P \gets f^*(\vec{x})$\;
    \KwRet{P}
\end{algorithm2e}

\subsection{\textbf{Approach 2}: Regression with Contextual Information}
\label{sub-sec:regression-context}

This approach differs from the previous one since it makes use of control-flow information in order to add contextual knowledge. The basic idea consists of adding a limited set of features able to encapsulate the control-flow path followed by a partial trace. We chose as control-flow model a transition system because it generally represents a good trade-off between expressivity and compactness. In Section~\ref{sub-sec:transition-system}, we showed how to construct a labeled transition system $\textit{TS} = (S, E, T)$ starting from an event log $L$. Now, we have to transform the TS into a series of features and encoding it into a proper form applicable to the $\epsilon$-SVR algorithm. As for the literal attributes, we use the one-hot encoding: the set $S \setminus S^\textit{start}$ is treated as a literal domain, where the possible values are the states $s \in S$ excluding the initial state because a non-empty trace always maps onto a state not included in $S^\textit{start}$. So, we enumerate the states, $s_1, s_2, \dots, s_n \in S \setminus S^\textit{start}$, and we map a state $s_i$ into a vector $\vec{v} \in \{0, 1\}^n$ by setting to 1 the $i$-th component in the n-dimensional vector, and to 0 all the others. For example, given the states set $S \setminus S^\textit{start} = \{s_1, s_2, s_3, s_4\}$ we encode the state $s_3$ onto $\vec{v} \in \{0, 1\}^4$ such that $\vec{v} = [0, 0, 1, 0]$.

\subsubsection{Non-fitting Traces}

As for the previous approach, even this one is able to handle non-fitting traces. However, without any adjustment, the prediction is calculated overlooking the control-flow. Indeed, using the encoding described above, if $f^\textit{state}(\sigma) \notin S$ then the corresponding vector would be null (i.e., $\vec{v} = [0, 0, \dots, 0]$). We cope this problem by mapping the \emph{non-compliant} state $s = f^\textit{state}(\sigma) \notin S$, onto a set of lawful states $s_i \in S$. The idea is to associate the non-fitting trace with states that are, within some degree, similar. Then the vector $\vec{v}$ will contain for each state the normalized similarity value.
It is very important to define a thoughtful similarity function: we assume that two states are similar if their representations are similar. In particular, since we are focusing on control-flow, we use as event representation function something like $f^\textit{event}(e) = \pi_\mathcal{A}(e)$. This implies that abstractions are aggregate representations of a set of activities. 

Let us define a similarity function for each abstraction considered in Section~\ref{sec:background} (i.e., set, bag and sequence):

\begin{definition}[Set Similarity Function]
Given two sets $x_1, x_2 \subseteq \mathcal{X}$, with $\mathcal{X}$ the set of all possible values, we define the similarity function\\ $f^\textit{sim}_{\textit{set}} \in 2^\mathcal{X} \times 2^\mathcal{X} \to [0, 1]$ as the Jaccard similarity \cite{Retrieval2008}. Formally:
$$
    f^\textit{sim}_{\textit{set}}(x_1, x_2) = \frac{|x_1 \cap x_2|}{|x_1 \cup x_2|}
$$
\end{definition}

\begin{definition}[Bag Similarity Function]
Given two multi-sets over a root set $\mathcal{X}$, $x_1, x_2 \in \mathbb{B}(\mathcal{X})$, we define the similarity function $f^\textit{sim}_{\textit{bag}} \in \mathbb{B}(\mathcal{X}) \times \mathbb{B}(\mathcal{X}) \to [0, 1]$ as the Jaccard similarity \cite{Retrieval2008}. Formally:
$$
    f^\textit{sim}_{\textit{bag}}(x_1, x_2) = \frac{\#\left(x_1 \nplus x_2\right)}{\#\left(x_1 \cup x_2\right)}
$$
\end{definition}

\begin{definition}[List Similarity Function]
Given two finite sequences over $\mathcal{X}$, $x_1, x_2 \in \mathbb{S}(\mathcal{X})$, we define the similarity function $f^\textit{sim}_{\textit{list}} \in \mathbb{S}(\mathcal{X}) \times \mathbb{S}(\mathcal{X}) \to [0, 1] \subset \mathbb{R}$ as the Damerau-Levenhstein similarity~\cite{Damerau1964}. Formally:
$$
    f^\textit{sim}_{\textit{list}}(x_1, x_2) = 1 - \frac{f^\textit{dist}_\textit{D-L}(x_1, x_2)}{\max(|x_1|, |x_2|)} 
$$
\end{definition}
The Damerau-Levenshtein distance ($f^\textit{dist}_\textit{D-L}$) is a distance between two string. It is calculated by counting the minimum number of edit operations needed to transform a string into the other. The set of possible edit operations takes into account by this metric are:
\begin{itemize}
    \item \textbf{insertion} of one character, e.g., \emph{ac} becomes \emph{abc} with the insertion of \emph{b};
    \item \textbf{deletion} of one character, e.g., \emph{abc} becomes \emph{ac} with the deletion of \emph{b};
    \item \textbf{substitution} of a character, e.g., \emph{ab} becomes \emph{ac} with the substitution of \emph{b} with \emph{c};
    \item \textbf{transposition} of two characters, e.g., \emph{abc} becomes \emph{acb} with the transposition of \emph{b} and \emph{c}.
\end{itemize}
Since this metric works over strings, we need to convert a sequence of event representations into a string. To do this, we simply map each event onto a character.\\

On the basis of the abstraction used for constructing the transition system, the corresponding similarity function is chosen. Every time a non-fitting trace comes into play, its representation is compared with all the state representations of the TS (excluding the initial state).
So, given a transition system $\textit{TS} = (S, E, T)$ (created using $f^\textit{event}$ and $f^\textit{state}$), a similarity function $f^\textit{sim}$ and a trace $\sigma$ (such that $s' = f^\textit{state}(\sigma) \notin S$) $f^\textit{sim}(s, s')$ is computed for each state $s \in S \setminus S^\textit{start}$.
After that each similarity value is normalized and finally put into the vector $\vec{v}$. We will call this kind of TS \textit{similarity-based} transition system.

Formally, given $s' \notin S$ and a $s_i \in S \setminus S^\textit{start}$ the corresponding $i$-th component of the resulting vector $\vec{v}$ will contain the following value:
$$
	\frac{f^\textit{sim}(s', s_i)}{\sum\limits_{s \in S \setminus S^\textit{start}} f^\textit{sim}(s', s)}.
$$

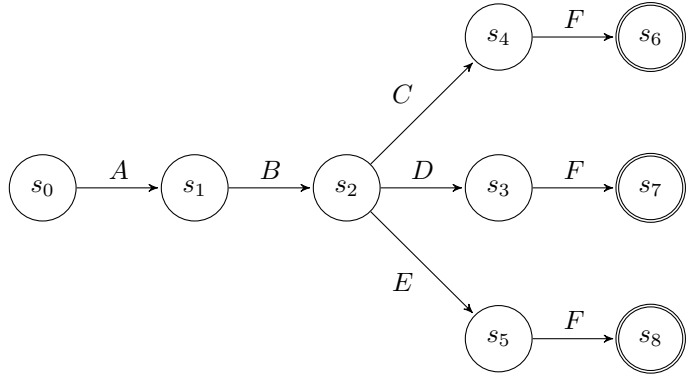
\begin{figure}
\centering
\begin{tikzpicture}[>=stealth',shorten >=1pt,auto,node distance=2cm]
    \node[state] (o)                {${s_0}$};
    \node[state] (A) [right of = o] {${s_1}$};
    \node[state] (B) [right of = A] {${s_2}$};
    \node[state] (D) [right of = B] {${s_3}$};
    \node[state] (C) [above of = D] {${s_4}$};
    \node[state] (E) [below of = D] {${s_5}$};
    \node[state,accepting] (F1) [right of = C] {${s_6}$};
    \node[state,accepting] (F2) [right of = D] {${s_7}$};
    \node[state,accepting] (F3) [right of = E] {${s_8}$};
    \path[->](o) edge node [align=center]{$A$} (A) 
             (A) edge node [align=center]{$B$} (B)
             (B) edge node [align=center]{$C$} (C)
             (B) edge node [align=center]{$D$} (D)
             (B) edge [below left] node [align=center]{$E$} (E)
             (C) edge node [align=center]{$F$} (F1)
             (D) edge node [align=center]{$F$} (F2)
             (E) edge node [align=center]{$F$} (F3);
\end{tikzpicture}
\caption{Example of a transition system extracted from a log containing three trace types $\langle A, B, C, F\rangle$, $\langle A, B, D, F\rangle$ and $\langle A, B, E, F\rangle$, with $f^\textit{event}(e) = \pi_\mathcal{A}(e)$ and $f^\textit{state}(\sigma) = \{f^\textit{event}(e) \mid e \in \sigma \}$. The state $s_0$ is the initial state, while $s_6, s_7, s_8$ are the accepting (i.e., final) states. Each transition is labeled by the corresponding event representation value.\label{fig:ts-no-hor}}
\end{figure}

Let us show a non-fitting trace management example: consider the transition system (i.e., $\textit{TS} = (S, E, T)$) in Fig.~\ref{fig:ts-no-hor} constructed using as state representation function the set abstraction (i.e., $f^\textit{state}(\sigma) = \{f^\textit{event}(e) \mid e \in \sigma \}$). In this example the state representation functions are: $s_0 = \{\}$, $s_1 = \{A\}$, $s_2 = \{A, B\}$, $s_3 = \{A, B, C\}$, $s_4 = \{A, B, D\}$, $s_5 = \{A, B, E\}$, $s_6 = \{A, B, C, F\}$, $s_7 = \{A, B, D, F\}$ and $s_8 = \{A, B, E, F\}$.
Given the non-fitting trace $\sigma' = \langle A, D \rangle$ we calculate the similarity of $s' = f^\textit{state}(\sigma') = \{A, D\}$ with every $s \in S \setminus \{s_0\}$ using $f^\textit{sim}_\textit{set}$:
\begin{align*}
	&f^\textit{sim}_\textit{set}(s', s_1) = \frac{1}{2} = 0.5, \:
	&f^\textit{sim}_\textit{set}(s', s_2) = \frac{1}{3} = 0.\bar{3}, \\
	&f^\textit{sim}_\textit{set}(s', s_3) = \frac{1}{4} = 0.25, \:
	&f^\textit{sim}_\textit{set}(s', s_4) = \frac{2}{3} = 0.\bar{6}, \\
	&f^\textit{sim}_\textit{set}(s', s_5) = \frac{1}{4} = 0.25, \:
	&f^\textit{sim}_\textit{set}(s', s_6) = \frac{1}{5} = 0.2,\\
	&f^\textit{sim}_\textit{set}(s', s_7) = \frac{2}{4} = 0.5, \:
	&f^\textit{sim}_\textit{set}(s', s_8) = \frac{1}{5} = 0.2.
\end{align*}
Then we normalize each value with the summation:
$$
	\sum_{s_i \in S \setminus \{s_0\}} f^\textit{sim}_\textit{set}(s', s_i) = 2.4
$$
obtaining the final vector:
\begin{align*}
	\vec{v} &= \left[\frac{0.5}{2.4}, \frac{0.\bar{3}}{2.4}, \frac{0.25}{2.4}, \frac{0.\bar{6}}{2.4}, \frac{0.25}{2.4}, \frac{0.2}{2.4}, \frac{0.5}{2.4}, \frac{0.2}{2.4}\right] \\
	&= \left[0.208\bar{3}, 0.13\bar{8}, 0.1041\bar{6}, 0.2\bar{7}, 0.1041\bar{6}, 0.08\bar{3}, 0.208\bar{3}, 0.08\bar{3}\right]
\end{align*}

\subsubsection{Training}
\label{sub-sub-sec:regression-context-training}

The training phase of this method is almost the same of the preceding one. The main difference lays on the introduction of a new derived feature to the training set. This can be done by making some minor changes to the Algorithm~\ref{alg:training-construction}. 
Since we assume the construction of $\textit{TS}$, we need it as input along with the state representation function $f^{\textit{state}}$ and the similarity function $f^\textit{sim}$. We calculate the state associated to each partial traces and we encode it into a one-hot vector or into the normalized similarity vector if it is a non-fitting trace. Finally, we construct the rest of the training instances as in Alg.~\ref{alg:training-construction}. Algorithm~\ref{alg:non-fitting} show the non-fitting trace management routine.

\begin{algorithm2e}
    \caption{Non-fitting trace encoder\label{alg:non-fitting}}
    \SetKwComment{tcp}{$\triangleright\ $}
	\DontPrintSemicolon
	\KwIn{$\sigma$: non-fitting trace, $S$: states set, $f^{\textit{state}}$: state representation function, $f^\textit{sim}$: similarity function}
	\KwOut{$\textit{Tr}$: training set}
	\DontPrintSemicolon
	\BlankLine
	$s \gets f^\textit{state}(\sigma)$ \;
	$\vec{v} \gets []$ \;
	$\textit{den} \gets 0$ \; 
	\For(\tcp*[f]{calculate the normalization factor}){$s_i \in S$} {\label{alg-line:first-loop}
    	$\textit{den} \gets \textit{den} + f^\textit{sim}(s_i, s)$\;
    }
    \For(\tcp*[f]{append the normalized vector}){$s_i \in S$} {\label{alg-line:sec-loop}
    	$c \gets \frac{f^\textit{sim}(s_i, s)}{\textit{den}}$ \;
		$\vec{v} \gets \vec{v} \: || \: [c]$
    }    		
    \KwRet{$\vec{v}$}
\end{algorithm2e}

First loop (line~\ref{alg-line:first-loop}) calculates the normalization factor, while the second loop (line~\ref{alg-line:sec-loop}) creates the vector.

\subsubsection{Prediction}
\label{sub-sub-sec:regression-context-prediction}

In this phase, as for the Simple Regression approach, the $\epsilon$-SVR model created in the previous step is used to forecast the remaining time of running process instances. The novelty of the method just described consists of the resulting model, which is obtained from the training phase. The introduction of contextual information, generally, leads to a different optimization problem and consequently to a different final model. The only changes to make in Alg.~\ref{alg:svr-prediction} are adding $f^\textit{state}$ as input, and substitute the right side of line~\ref{alg-line:first} with the one-hot (or the non-fitting) encoding of $f^\textit{state}(\sigma_p)$.

\subsection{\textbf{Approach 3}: Data-aware Transition System (DATS)}
\label{sub-sec:ensemble}

The approach presented in this section is a refinement of \cite{Polato2014} which exploits the same idea described in \cite{VanderAalst2009}. Let us recall the main characteristics of the latter method. In their work van der Aalst et al. introduced the concept of annotated transition system: each state of the transition system is ``decorated'' with predictive information, called \emph{measurements}. Since we want to predict the remaining time, a possible set of measurements collected in a state might be the remaining time starting from the state itself. Formally, in \cite{VanderAalst2009} a measurement is defined as:

\begin{definition}[Measurement]
A measurement function $f^\textit{measure}$ is a function that, given a trace $\sigma$ and en event index $i \in [1, 2, \dots, |\sigma|]$ produces some measurement. Formally, $f^\textit{measure} \in \Sigma \times \mathbb{N} \to \mathcal{M}$, where $\mathcal{M}$ is the set of possible measurement values (e.g., remaining time).
\end{definition}

In \cite{VanderAalst2009} different kinds of measurements are proposed. Once the suitable measurement is chosen, an annotated transition system is created according to the event log:

\begin{definition}[Annotated transition system]
Let $L$ be an event log and $\textit{TS} = (S, E, T)$ a labeled transition system constructed over $L$ using the representation functions $f^\textit{event}$ and $f^\textit{state}$. Given a particular measurement function $f^\textit{measure} : \Sigma \times \mathbb{N} \to \mathcal{M}$, we define an annotation $A \in S \to \mathbb{B}(\mathcal{M})$, such that $\forall s \in S$:
$$
	A(s) = \biguplus_{\sigma \in L} \biguplus_{\stackrel{1\leq k \leq |\sigma|}{s = f^\textit{state}(\sigma^k)}} f^\textit{measure}(\sigma, k)
$$
An annotated transition system is the tuple $(S, E, T, A)$.
\end{definition}

Since we are facing the remaining time prediction problem, our $f^\textit{measure}$ function coincides with the function $rem$ defined in Section~\ref{sec:prediction}. The last step consists of the definition of a prediction function $\in \mathbb{B}(\mathcal{M}) \to \mathcal{M}$, such that given a multiset of measurements produces some prediction, e.g., the average or the median value. So, in the operative setting we have an annotated transition system $(S, E, T, A)$ constructed over $L$ and a prediction function $\mathcal{P} : \mathbb{B}(\mathcal{M}) \to \mathcal{M}$. Using these tools a prediction is made in a straightforward way: given a partial trace $\sigma_p$ observed so far, such that $f^\textit{state}(\sigma_p) = s$, the prediction value is $\mathcal{P}(A(s))$. It is worth to notice that the prediction is calculated using merely control-flow (i.e., transition system) and temporal (i.e., remaining time) information.\\

The main difference introduced by our approach is the addition of classifiers and regressors, which take advantage of additional attributes, as annotations. Let us give a brief overview of this approach. 
As in \cite{VanderAalst2009}, we start with the transition system construction and then we enrich each state with a \Naive~Bayes classifier (see Section~\ref{sub-sec:bayes}) and each transitions with a Support Vector Regressor (see Section~\ref{sub-sec:svr}) trained with historical data considering all attributes. 
The introduction of these two machine learning models is based on the intuition that in a state $s$ \Naive~Bayes estimates the probability of transition from $s$ to $s' \in s\bullet$, while SVR predicts the remaining time if the next state will be $s'$.

\begin{figure}
\centering
\begin{tikzpicture}[>=stealth',shorten >=1pt,auto,node distance=2.7cm]
    \node (o){};
    \node[state] (B) [below right of = o] {${s_2}_{\{B\}}$};
    \node[state] (D) [below of = B] {${s_4}_{\{D\}}$};
    \node[state] (C) [left of = D] {${s_3}_{\{C\}}$};
    \node[state] (E) [right of = D] {${s_5}_{\{E\}}$};
    \path[->](o) edge [dashed, bend left] node {}(B)
             (B) edge [above left] node [align=center]{$0.6$} (C)
             (B) edge node {$0.1$} (D)
             (B) edge node {$0.3$} (E);
\end{tikzpicture}
\caption{A state annotated with a \Naive~Bayes classifier. The probability of going from state $s_2$ to any of its exiting states is reported next to the corresponding edge. These probabilities are the result of the Na\"ive Bayes classifier, which takes as input the data attributes of the running case.\label{fig:naive}}
\end{figure}
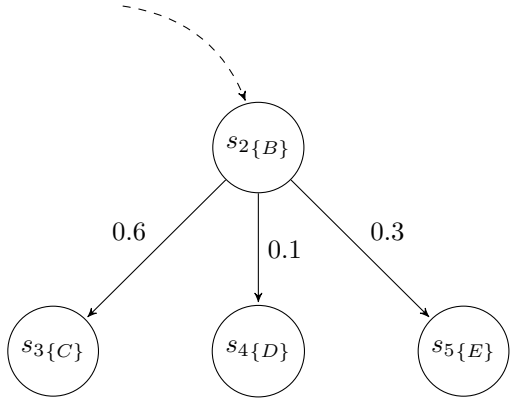

Figure~\ref{fig:naive} proposes an example of state ($s_2$) annotated with a \Naive~Bayes classifier. In this state, the Na\"ive Bayes classifier gets the probabilities to reach each exiting state: probabilities to reach states $s_3, s_4$ and $s_5$ are respectively 0.6, 0.1 and 0.3. Such values are used to weigh the remaining time values obtained from the support vector regressors. In Figure~\ref{fig:naive} the state $s_2$ corresponds to the current partial trace $\sigma'$. From each outgoing state (i.e., $s_3, s_4, s_5$) the remaining time is estimated using the SVR associated to the incoming transition (i.e., $s_2 \to s_3$, $s_2 \to s_4$ and $s_2 \to s_5$). 

\begin{figure}
\centering
\begin{tikzpicture}[>=stealth',shorten >=1pt,auto,node distance=3.2cm]
    \node (o){};
    \node[state] (B) [below right of = o] {${s_2}_{\{B\}}$};
    \node[state] (D) [below of = B] {${s_4}_{\{D\}}$};
    \node[state] (C) [left of = D] {${s_3}_{\{C\}}$};
    \node[state] (E) [right of = D] {${s_5}_{\{E\}}$};
    \path[->](o) edge [dashed, bend left] node {}(B)
             (B) edge [bend right, above left] node [align=center]{\footnotesize Remaining: 2 h} (C)
             (B) edge node [align=center]{\footnotesize Remaining: 3 h} (D)
             (B) edge [bend left, above right] node [align=center]{\footnotesize Remaining: 1 h} (E);
\end{tikzpicture}
\caption{Example of a Support vector regressor application. The estimated remaining times are suggest by the label \emph{Remaining}.\label{fig:svr}}
\end{figure}
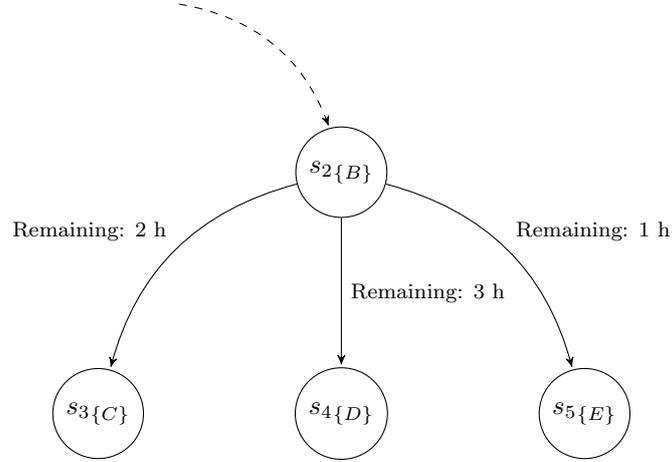

These estimations are multiplied by the probability values obtained from the NB classifiers, and finally summed together in order to compute a weighted average over all the possible continuation from $s$.
Figure~\ref{fig:svr} presents an example of remaining time estimation for each outgoing state. In this example the estimated remaining time, starting from the state $s_3$, is 2 hours, while the average duration of the transition $s_2 \to s_3$ is 20 minutes. 

Formally, let $\hat{p}_{s'}$ be the \Naive~Bayes estimated probability to reach state $s' \in s\bullet$ from state $s$, and $\hat{\tau}_{s \to s'}$ the estimated remaining time returned by the SVR associated to transition $s \to s'$. 
Then, given the state $s$ reached after observing a (partial) trace $\sigma$, the prediction returned by the annotated transition system is $\sum_{s' \in s\bullet} (\hat{p}_{s'} \cdot \hat{\tau}_{s \to s'})$.

Let us now define the annotations used to decorate the transition system. 

\begin{definition}[\Naive~Bayes Annotation]
Let $\textit{TS}$ be a labeled transition system, obtained from an event log $L$, based on an event representation function $f^{\textit{event}}$ and a state representation function $f^{\textit{state}}$. Let's call $k$ the size of the $\gamma^*(\sigma)$ vector calculated for traces $\sigma \in L$.
A \emph{\Naive~Bayes Annotation} is a function $\mathit{NB} : S \times \mathbb{R}^k \times S \rightarrow [0, 1] \subset \mathbb{R}$, which, given two states $s_i, s_j \in S$ and a data attribute vector $\vec{x} \in \mathbb{R}^k$, returns the probability to reach the state $s_j$ starting from $s_i$ through a single transition.
\end{definition}

\begin{definition}[SVR Annotation]
Let $\textit{TS}$ be a labeled transition system, obtained from an event log $L$, based on an event representation function $f^{\textit{event}}$ and a state representation function $f^{\textit{state}}$.
An \emph{SVR Annotation} is a function $R : T \times \mathbb{R}^k \rightarrow \mathbb{R}$, such that, given a transition $t \in T$ and a data attribute vector $\vec{x} \in \mathbb{R}^k$, it applies Support Vector Regression to produce an estimation of the remaining time. 
\end{definition}

Using these annotations, we define a predictor transition system as follows:

\begin{definition}[Predictor TS]
Let \mbox{$\textit{TS} = (S, E, T)$} be a labeled transition system, obtained from an event log $L$, based on an event representation function $f^{\textit{event}}$ and a state representation function $f^{\textit{state}}$. A \emph{predictor transition system} is a tuple $\textit{PTS} = (S, E, T, \textit{NB}, R)$ where $\textit{NB}$, $R$ are respectively a \Naive~Bayes and a SVR annotation, based on the event log $L$ and the transition system $\textit{TS}$.
\end{definition}

\subsubsection{Training}
In this section, we are going to describe how to construct a predictor transition system. 
Algorithm~\ref{alg:pts-construction} shows the construction procedure.

\begin{algorithm2e}[H]
    \caption{Construction of a Predictor Transition System}
    \label{alg:pts-construction}
    \SetKwComment{tcp}{$\triangleright\ $}
    \DontPrintSemicolon
	\KwIn{$L$: event log; $\textit{TS} = (S, E, T)$: labeled transition system}
	\KwOut{$T'$: predictor transition system}
    \DontPrintSemicolon
	\BlankLine
	\ForEach{$t \in T$}{ \label{alg-line:first-loop-start}
    	$\textit{svr}[t] = \emptyset$ \tcp*[r]{training set for $t$} 
    } \label{alg-line:first-loop-end}
    \BlankLine
    \ForEach{$\sigma_c \in L$} { \label{alg-line:sec-loop-start}
		\For{$i \gets 1\ \mbox{\bf to}\  |\sigma_c| - 1$} {
    		$s \gets f^{\textit{\textit{state}}}(\sigma_c^i)$ \; \label{alg-line:tr-1}
    		$s' \gets f^{\textit{\textit{state}}}(\sigma_c^{i+1})$ \; \label{alg-line:tr-2}
    		$e \gets f^{\textit{event}}({\sigma_c(i+1)})$ \; \label{alg-line:tr-3}
		    $t \gets (s, e, s')$ \; \label{alg-line:tr-4}
		    $\vec{x} \gets \gamma^*(\sigma_c^i)$ \; \label{alg-line:tr-5}
		    $y \gets \mathit{rem}(\sigma_c, i)$ \; \label{alg-line:tr-6}
		    $\textit{svr}[t] \gets \textit{svr}[t] \cup (\vec{x}, y)$ \; \label{alg-line:tr-7}
		    \BlankLine
		    \If{$|s\bullet| \geq 2$}{ \label{alg-line:nb-start}
				Update \textit{NB} for state $s$ with instance $(\vec{x}, s')$ \label{alg-line:nb}
			}\label{alg-line:nb-end}
    	}
    }\label{alg-line:sec-loop-end}
	\BlankLine
	\ForEach{$t \in T$} { \label{alg-line:svr-start}
    	Train SVR ($R$) for transition $t$ with training set $\textit{svr}[t]$ \;
    }\label{alg-line:svr-end}
    \BlankLine
	$T' \gets (S, E, T, \mathit{NB}, R)$ \;
    \KwRet{$T'$}
\end{algorithm2e}
The first loop (line~\ref{alg-line:first-loop-start}) initializes all the training sets for the $\epsilon$-SVR model to the empty set, while the second loop (lines~\ref{alg-line:sec-loop-start}-\ref{alg-line:sec-loop-end}) creates the training sets and updates the NB classifiers. In particular, lines~\ref{alg-line:tr-1}-\ref{alg-line:tr-7} construct the training instances by extracting the additional data from the partial traces and calculating the remaining time. Being possible to build a NB model incrementally, this is done in line~\ref{alg-line:nb}. Last loop (lines~\ref{alg-line:svr-start}-\ref{alg-line:svr-end}) trains the $\epsilon$-SVR models using the training sets built previously. 

\subsubsection{Prediction}
In this section, we describe how to predict the remaining time for a running case using a predictor transition system constructed with Alg.~\ref{alg:pts-construction}. Algorithm~\ref{alg:prediction} shows the prediction procedure.

\begin{algorithm2e}
    \caption{Remaining time prediction for a running case}
    \label{alg:prediction}
	\DontPrintSemicolon
	\KwIn{$\sigma_p$: (partial) trace; $\textit{PTS} = (S, E, T, \textit{NB}, R)$: predictor transition system}
	\KwOut{$P$: remaining time prediction}
	\BlankLine
    $P \gets 0$ \;
    $s \gets f^{\textit{state}}(\sigma_p)$ \;
    $\vec{x} \gets \gamma^*(\sigma_p)$ \;
    \BlankLine
    \eIf {$|T_{s}| \geq 2$} {
        \ForEach{$t = (s, e, s') \in T_{s}$}{
            $P \gets P + \mathit{NB}(s, \vec{x}, s') \cdot R(t, \vec{x})$ \;
        }
    }{
        $s' \gets f^{\textit{state}}(\sigma_p^{|\sigma_p|-1})$ \;
        $t' \gets (s', e, s) \in T_{s'}$ \;
        $P \gets R(t', \vec{x})$ \;
    }
    \KwRet{$P$}
\end{algorithm2e}

The algorithm simply applies the formula, seen at the beginning of this section, $\sum_{s' \in s\bullet} (\hat{p}_{s'} \cdot \hat{\tau}_{s \to s'})$. Each $\hat{p}_{s'}$ is a value produced by the application of the NB classifiers and $\hat{\tau}_{s \to s'}$ by the $\epsilon$-SVRs. Specifically, let $s = f^\textit{state}(\sigma_p)$, then $\hat{p}_{s'} = \textit{NB}(s, \gamma^*(\sigma), s')$ and $\hat{\tau}_{s \to s'} = R(s \to s', \gamma^*(\sigma))$, for all $s \in s\bullet$. A core difference w.r.t. \cite{Polato2014} is the absence of the expected sojourn time (on the current state): in this revised version this information is implicitly embedded inside the $\epsilon$-SVR and hence can be removed from the formulation.

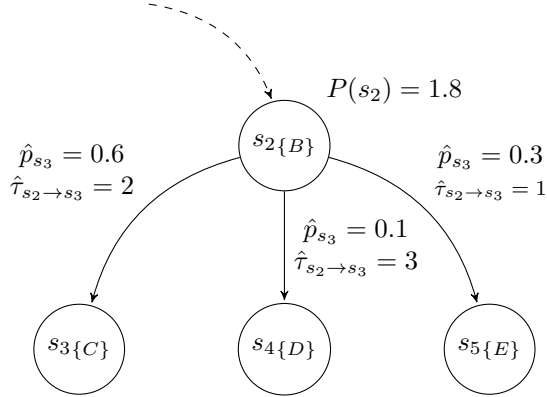
\begin{figure}[ht]
\centering
\begin{tikzpicture}[>=stealth',shorten >=1pt,auto,node distance=2.7cm]
    \node (o){};
    \node[state] (B) [below right of = o,label=above right:{$P(s_2)=1.8$}] {${s_2}_{\{B\}}$};
    \node[state] (D) [below of = B] {${s_4}_{\{D\}}$};
    \node[state] (C) [left of = D] {${s_3}_{\{C\}}$};
    \node[state] (E) [right of = D] {${s_5}_{\{E\}}$};
    \path[->](o) edge [dashed, bend left] node {}(B)
             (B) edge [bend right, above left] node [align=center]{$\hat{p}_{s_3} = 0.6$\\  $\hat{\tau}_{s_2 \to s_3} = 2$} (C)
             (B) edge node [align=center]{$\hat{p}_{s_3} = 0.1$\\  $\hat{\tau}_{s_2 \to s_3} = 3$} (D)
             (B) edge [bend left, above right] node [align=center]{$\hat{p}_{s_3} = 0.3$\\ \footnotesize $\hat{\tau}_{s_2 \to s_3} = 1$} (E);
\end{tikzpicture}
\caption{Example of a prediction calculated by a predictor transition system: \mbox{$P(s_2) = 0.6 \cdot 2 + 0.1 \cdot 3 + 0.3 \cdot 1 = 1.8$}.\label{fig:pts}}
\end{figure}

Figure~\ref{fig:pts} puts together the two representations depicted in Section~\ref{sub-sec:ensemble}: it shows an example of NB and SVR application for a partial trace $\sigma = \langle A, B\rangle$ (see event log in Table~\ref{tb:log}). The remaining time prediction, in hours, for this example is: 
$$
	P(s_2) = \sum_{s' \in \{s_3, s_4, s_5\}} (\hat{p}_{s'} \cdot \hat{\tau}_{s_2 \to s'}) = 0.6 \cdot 2 + 0.1 \cdot 3 + 0.3 \cdot 1 = 1.8.
$$

Please note that, given a predictor transition system (result of the learning procedure), the computation of the prediction requires constant number of operations which, in the worst case corresponds to the size of the largest set $s\bullet$ of the transition system. This property allows the application of the approach in on-line settings \cite{burattin15}, where each event is allowed to trigger only a constant number of operations.

\subsubsection{Future Path Prediction}
\label{sec:path}
The model we use in this last approach can also be exploit in order to predict, for a running case, which is the most likely sequence of activities until the end of the case. Let us take for example the situation depicted in Fig.~\ref{fig:naive} which is a fragment of the TS in Fig.~\ref{fig:ts}: the most likely sequence of states starting from $s_{2\{B\}}$ is $s_{2\{B\}} \rightarrow s_{3\{C\}} \rightarrow s_{6\{F\}}$, $s_{6\{F\}}$ is an accepting node and so the process instance is complete, with a probability equals to $0.6 * 1 = 0.6$. Note that the transition between $s_{3\{C\}}$ and $s_{6\{F\}}$ has a probability equals to 1 because is the only way the process can proceed. In general, the sequence can go through many split states in which the transition probabilities are $< 1$ and find the full sequence with the higher probability is not a trivial task.

We face this problem as a shortest path problem in which the goal is to find a path between to vertices of a graph such that the sum of weights of its constituent edges is minimized. Specifically, let us consider the transition system as a directed graph: we would like to find the shortest path between the current node and an accepting node. In order to leverage this idea, we need to define a suitable distance measure (i.e., cost) between nodes. 

Given a possible sequence of activities (in a Predictor TS) between the node $s_1$ to $s_n$, i.e., $s_1, s_2, \dots, s_n$,  with the corresponding transition probabilities (obtained using the NB annotations), i.e., $p_i \forall s_i \rightarrow s_{i+1}, 1 \leq i < n$, then the likelihood of such sequence is defined by $\prod_{1\leq i < n} p_i$.
Since probabilities have to be multiplied to get the likelihood of a sequence we cannot use it directly as edge cost because we need to follow the definition of the shortest path problem (i.e., edge cost have to be summed). However, we can exploit properties of the \textit{logarithm} function to transform probabilities into distance like values, in particular:
$$
\log(pq) = \log(p) + \log(q),
$$
and since $p$ are probabilities:
$$
\forall 0 \leq p \leq 1 \in \mathbb{R} , \log(p) \leq 0,
$$
where low values of $log(p)$ mean that the transition probability is low, or, from a graph view point, the distance between the nodes is high. Using this idea, we can use as edge cost the opposite of the logarithm of the transition probability. Note that this logarithm transformation works as we desire: probabilities close to 0 (i.e., rare occurrences) correspond to high costs while probabilities close to 1 (i.e., very likely) correspond to costs almost null. 
Using the just mentioned transformation we can construct a graph corresponding to the TS where the shortest path problem can be solved applying a best first search. So, from a computational point of view the prediction of the activity sequence has a cost which is quadratic in the number of nodes (i.e., states of the TS).

\section{Implementation}
\label{sec:implementation}

These techniques have been implemented for the ProM framework~\cite{Verbeek2009}. To mine the transition systems, we rely on the miner's implementation available inside the framework. Na\"ive Bayes Classification and the Support Vector Regression are performed using the implementation in the Weka framework~\cite{Hall2009}. In particular, for SVR we used the SMO (Sequential Minimal Optimization) implementation provided by the framework.

\begin{figure}
	\includegraphics[width=\textwidth]{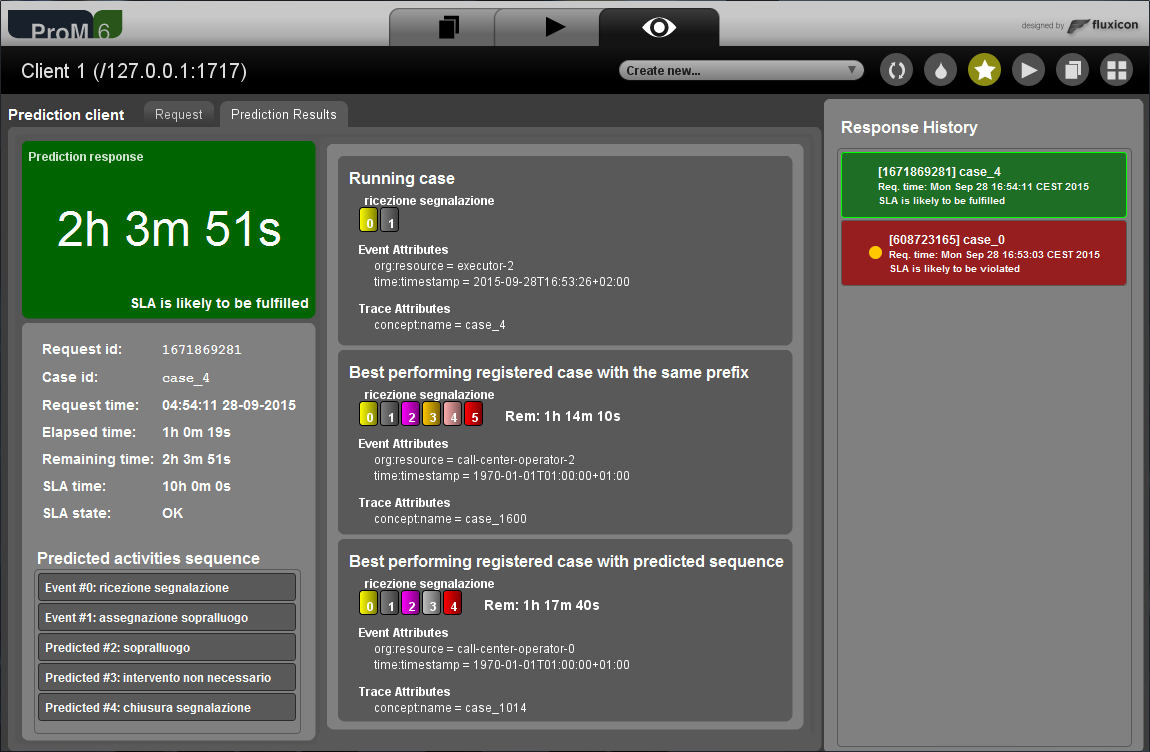}
	\caption{ Screenshot of the ProM plugin that has been developed, and that can be used to manually query the prediction model.}
	\label{fig:screenshot}
\end{figure}

The main ProM plugin we developed is capable of building a prediction model according to the methods described earlier on this paper. Once the prediction model has been created, another ProM plugin exposes an on-line service which can be queried for predictions. This online querying layer uses JSON\footnote{JSON stands for ``JavaScript Object Notation''. More information can be retrieved from \url{http://www.json.org/}.} as communication languages and, therefore, in principle, any information system, implementing the query protocol, can embed the prediction features that we provide.
Moreover, apart from these prediction functionalities, we also built another plugin which can be used to manually query the model. A screenshot of this plugin is reported in Figure~\ref{fig:screenshot}.
The plugin allows users to ask for prediction on a specific running instance, and the user can also specify a possible deadline for it. In case the prediction does not fulfill the deadline, an alarm is raised.

The figure shows how a prediction is visualized to the client-side plugin. On the left, there are all the information regarding the remaining time prediction (upper left corner) and also the activity sequence prediction (bottom left). The predicted time is highlighted in green when the provided deadline seems to be fulfilled in red otherwise. On the right side, there is the history of the queries, while the middle section shows two historical traces which have been the best performances. Specifically, these traces have all the same initial activities as the running case (shown as the first trace), but they are the ones which have the shortest remaining time. What differ is that one represent the fastest trace in the entire log regardless of the sequence of activities, while the second one (third in figure) is the fastest one which follows the predicted sequence of future activities. 
All these aim to support as much as possible the user in her decision making process.

Due to the simplicity of the prediction procedures, the time for computing the forecast is negligible, thus paving the way to the adoption of our approach into intensive and on-line scenarios.

\section{Results}
The experiments reported in this section aim to assess how the techniques leveraging on the similarity-based transition system give more accurate predictions for variants of process executions that have not been observed in the training set. The comparison is made with respect to the specialization of the technique reported in van der Aalst et al.~\cite{VanderAalst2009} that has been discussed in Section~\ref{sec:related-work}.
We also aimed to compare our approach versus the techniques proposed in~\cite{Folino2012,Folino2013}. Unfortunately, up to this date, neither an implementation of these approaches nor the event logs used in the papers are publicly available. Of course, it is not possible to compare results obtained through different event logs, as the quality of a prediction heavily depends on the information available in the event log.
Work~\cite{Polato2014} shows that no much difference can be appreciated when polynomial or RBF is used as kernel type. Therefore, the experiments only makes use of the latter.

The comparison is made using two real-life case studies: the first concerns with a ticketing management process of the help desk of an Italian software company, and the second with the execution of process instances in an information system for the management of road-traffic fines by a local police office of an Italian municipality.

To measure and compare the accuracy, we used two indicators: the Mean Absolute Percentage Error (MAPE) and the Root Mean Square Percentage Error (RMSPE).
Let $n$ be the number of samples and let $A_i$ and $F_i$ be respectively the actual value and the predicted value for the $i$-th sample.
MAPE usually expresses the accuracy as a percentage:
\[
\textit{MAPE} = \frac{100\%}{n}\sum_{i=1}^n \left| \frac{A_i-F_i}{A_i}\right|,
\]
RMSPE is also defined as a percentage:
\[
\textit{RMSPE} = 100\% \sqrt{\frac{\sum_{i=1}^n (\frac{A_i-F_i}{A_i})^2}{n}}.
\]

\subsection{Case Study 1: Road Fines Log}

The first case study log concerns the execution of process instances in an information system for the management of road-traffic fines by a local police office of an Italian municipality. The management of road-traffic fines has to comply with Italian laws, which detail the precise work-flow. Usually, when a driver commits a violation, a policeman opens a new fine management and leaves a ticket on the car glass. The actual fine amount depends on the violation performed. Within 180 days, the fine notification must be sent to the offender.
The payment can occur in any moment, i.e.\ before or after that the fine notification is sent by post.
If the offender does not pay within 60 days since the reception of the fine notification, the fine doubles. If the offender never pays, eventually the fine is sent to a special agency for credit collection. In a number of circumstances, the fine can be dismissed. For instance, the policeman could make a typo when the fine form is filled in (e.g., the license plate number does not match the type of car/truck/motorbike). More importantly, the offender may think to have come under injustice and, hence, can appeal to a judge and/or the local prefecture. If the appeal is in favour to the offender, the fine is dismissed. Otherwise, the management is carried on as if he/she had never appealed.

We extracted from the information systems an event log that refers to executions that end with sending for credit collection, i.e.\ the offenders have not paid the fine in full. These event logs refer to non-overlapping periods in time and contains about 7300 log traces. 
With this set of experiments we want to show that all the methods presented here perform well under the assumption that in the training set there are all the possible behaviours of the process (the same assumption is done by van der Aalst et al. in \cite{VanderAalst2009}).
The experiments were performed using 5-fold cross validation. The SVR hyper-parameters have been tuned automatically using a grid search strategy. In particular we sought for the best combination of $C$, as penalty in the optimization problem, and $\gamma$ for the RBF kernel.
The transition system has been mined using different abstractions, namely set, multi-set and sequence with no limit. Since, in the event log, for $99\%$ of traces, every activity was performed at most once, and the process is almost linear, the multi-set and the sequence abstraction was not considered to perform experiments.

Table~\ref{tab:road-fines-ex} reports the results of the experiments. The acronym used in the tables mean: VDA is the approach presented in \cite{VanderAalst2009}, DATS is the data-aware transition system, SVR is the approach which use a simple support vector regression machine, and SVR+TS is the method which incorporates contextual information (via TS) in the training instances.
\begin{table}[h]
	\centering
	\begin{tabular}{r|cc}
		\toprule
		& \textbf{MAPE} & \textbf{RMSPE} \\
		\midrule
		\textbf{VDA*} & 5.89$\pm$0.14\% & 8.80$\pm$0.91\%  \\
		\textbf{DATS} & 2.82$\pm$0.10\% & \textbf{6.06$\pm$1.01\%}  \\
		\textbf{SVR} & 2.83$\pm$0.07\% & 6.72$\pm$1.40\%  \\
		\textbf{SVR+TS} & \textbf{2.77$\pm$0.09\%} & 6.58$\pm$1.41\%  \\ 
		\bottomrule
	\end{tabular}
	\caption{Road Fines log experiment results using 5-fold cross validation: (*) means that the approach is the baseline.}
	\label{tab:road-fines-ex}
\end{table}

Results show that every proposed approach outperform the baseline with similar results. However, it is worth to notice that the approaches based on the transition system, i.e., DATS and SVR+TS, achieve best performances on RMSPE and MAPE respectively. Readers can observe that VDA achieves better performance than in \cite{Polato2014}, this is due to the fact that the log used in these experiments is much bigger and hence the statistics of the methods have a stronger support. 

From these results, we can argue that the improvements with respect to the baseline are due to the introduction of the additional data in our models. In this log, we can also notice that the effect of considering the workflow (i.e., the transition system) is cramped in comparison to the data perspective. 

\subsection{Case Study 2: Help Desk Log}

The second case study log concerns the ticketing management process of the help desk of an Italian software company. In particular, this process consists of 14 activities: it starts with the insertion of a new ticket and then a seriousness level is applied. Then the ticket is managed by a resource and it is processed. When the problem is resolved it is closed and the process instance ends. This log has almost $4\,500$ instances with over $21\,000$ events. In this case the process is not linear, but it is well structured.
As in the previous case study, the experiments were performed using 5-fold cross validation and the SVR's hyper-parameters have been tuned automatically using a grid search strategy. Since the process it is more complex, we performed two kinds of experiments: 
\begin{itemize}
	\item we compared the performance of our approaches against the baseline assuming that in the training phase all the possible behaviours of the process are present at least once;
	\item we removed some variants from the training set in order to have completely new activity sequences in the test set, so the assumption is not valid anymore.
\end{itemize}
The second type of experiments aim to show that the approaches based on single SVR performed well even with noisy instances or with new events.\\

Tables~\ref{tab:ticket-ex-seq},~\ref{tab:ticket-ex-bag} and~\ref{tab:ticket-ex-set} show the results using the entire log with 5-fold cross validation. We performed tests using all the three abstractions (i.e., set, multi-set and sequence). With this dataset the usage of different set abstractions did not provide any remarkable difference. As for the previous case study, the best performing methods are DATS and SVR+TS, however the lowest MAPE and RMSPE are achieved by DATS using the set abstraction. Since the SVR approach had not good results (it is in line with VDA), the transition system information used by SVR+TS played a decisive role. The improvements with respect to the baseline are in average $14\%$ for the MAPE and $4\%$ for RMSPE. 

\begin{table}[h]
\centering
	\begin{tabular}{r|cc}
		\toprule
		& \multicolumn{2}{c}{\textbf{Sequence}} \\
		& \textbf{MAPE}	& \textbf{RMSPE}	\\
		\midrule
		\textbf{VDA*}	&29.94$\pm$0.59\%	&45.10$\pm$1.3\% \\
		\textbf{DATS}	&26.96$\pm$0.7\%	&43.76$\pm$0.91\% \\
		\textbf{SVR}	 &28.82$\pm$0.57\%	&47.78$\pm$0.83\% \\
		\textbf{SVR+TS}	 &\textbf{25.80$\pm$0.51\%}	&\textbf{41.90$\pm$0.98\%} \\
		\bottomrule
	\end{tabular}
\caption{Help Desk log experiment results using 5-fold cross validation with the sequence abstraction: (*) means that the approach is the baseline.}
\label{tab:ticket-ex-seq}
\end{table}

\begin{table}[h]
\centering
	\begin{tabular}{r|cc}
		\toprule
		& \multicolumn{2}{c}{\textbf{Multi-set}} \\
		& \textbf{MAPE}	& \textbf{RMSPE}\\
		\midrule
		\textbf{VDA*}	&29.58$\pm$0.35\%	&43.32$\pm$0.25\%	\\
		\textbf{DATS}	&26.54$\pm$0.51\%	&\textbf{42.16$\pm$0.61\%}\\
		\textbf{SVR} &28.03$\pm$1.1\%	&42.39$\pm$1.61\%	\\
		\textbf{SVR+TS}	&\textbf{25.84$\pm$0.55\%}	&42.70$\pm$1.01\%	\\
		\bottomrule
	\end{tabular}
\caption{Help Desk log experiment results using 5-fold cross validation with the multi-set abstraction: (*) means that the approach is the baseline.}
\label{tab:ticket-ex-bag}
\end{table}

\begin{table}[h]
\centering
	\begin{tabular}{r|cc}
		\toprule
		& \multicolumn{2}{c}{\textbf{Set}} \\
		& \textbf{MAPE}	& \textbf{RMSPE} \\
		\midrule
		\textbf{VDA*}	&29.94$\pm$0.36\%	&43.44$\pm$0.43\%	\\
		\textbf{DATS}	&\underline{\textbf{25.66$\pm$0.34\%}}	&\underline{\textbf{41.16$\pm$0.16\%}} \\
		\textbf{SVR}	 & 29.12$\pm$0.29\%	&47.46$\pm$3.1\% \\
		\textbf{SVR+TS}	&25.96$\pm$0.26\%	&42.76$\pm$1.41\% \\
		\bottomrule
	\end{tabular}
\caption{Help Desk log experiment results using 5-fold cross validation with the set abstraction: (*) means that the approach is the baseline. The \underline{underline} results are the overall best.}
\label{tab:ticket-ex-set}
\end{table}

From a workflow point of view, this case study has a more complex structure and we can see how the workflow information, via the TS, had different impact on the results. In this experiment the simple SVR failed in comparison with SVR+TS and the DATS methods, emphasizing the importance of the information brought by the TS.

\begin{table}[h]
\centering
\small
\begin{tabular}{r|cc|cc}
	\toprule
	& \multicolumn{2}{c|}{\textbf{Variants}} & \multicolumn{2}{c}{\textbf{Activity}} \\
	& \textbf{MAPE}	& \textbf{RMSPE}	& \textbf{MAPE}	& \textbf{RMSPE} \\
	\midrule
	\textbf{VDA*}	&41.26$\pm$1.11\%	&67.96$\pm$2.56\%	&41.28$\pm$0.27\%	&69.02$\pm$0.64\% \\
	\textbf{DATS}	&40.74$\pm$1.4\%	&67.50$\pm$2.21\%	&40.74$\pm$0.27\%	&69.02$\pm$0.42\% \\
	\textbf{SVR}	&41.48$\pm$1.2\%	&69.08$\pm$2.12\%	&41.00$\pm$0.25\%	&69.50$\pm$0.68\% \\
	\textbf{SVR+TS}	&\textbf{33.72$\pm$0.9\%}	&\textbf{54.44$\pm$2.6\%}	&\textbf{33.92$\pm$0.39\%}	&\textbf{55.84$\pm$0.77\%} \\
	\bottomrule
\end{tabular}
\caption{Help Desk log experiment (without some process variants) results using 5-fold cross validation: (*) means that the approach is the baseline.}
\label{tab:ticket-var-ex}
\end{table}

Table~\ref{tab:ticket-var-ex} shows the results with 5-fold cross validation on the log without some variants. In particular, column ``Variants'' shows results using training instances without half of the variants present in the starting event log. Column ``Activity'' instead, shows results removing from the training set all the traces with a specific activity, i.e., ``Waiting'', which were present in almost 25\% of the instances. Then the test phase uses the remaining part of the log, so inside the test there are completely new process behaviours which cause problems to VDA and DATS approaches.

Results show that in both cases, SVR+TS outperforms all the other approaches with a MAPE around $34\%$ and a RMSPE of $55\%$. The introduction of the similarity mechanism in SVR+TS makes this approach less sensible to the noise or change in the workflow, because it is able to mitigate the lack of the correct state using information from correlated ones. 

It is worth highlighting, that both VDA and DATS would not return any prediction in case of unseen activity' sequences. However, in order to be able to compare the approaches, we implemented a safety mechanism: if a trace does not map into a valid state of the transition system, the last event is removed and the mapping is redone. This process is repeated (see Alg.~\ref{alg:safety}) until obtaining a prefix that can be mapped or, viceversa, is empty (and is discarded), and in this case the average remaining time of the entire training is returned. This explains why SVR has not better performance than VDA and DATS. 

These experiments show the effectiveness of the SVR+TS approach and the importance of the similarity-based transition system.\\

\begin{algorithm2e}
    \caption{Safety mechanism}
    \label{alg:safety}
	\DontPrintSemicolon
	\KwIn{$\sigma_p$: (partial) trace; $\textit{TS} = (S, E, T)$: transition system}
	\KwOut{$s$: a valid state}
	\BlankLine
	$i \gets |\sigma_p|$\;
	$s \gets f^{\textit{state}}(\sigma_p^i)$ \;
	\While{$(s \notin S) \wedge (i > 0)$} {
		$i \gets i - 1$\;
    	$s \gets f^{\textit{state}}(\sigma_p^i)$ \;
    }
    \BlankLine
    \KwRet{$s$}
\end{algorithm2e}

\subsubsection{Future Sequence of Activities Prediction}

\begin{figure}[h!]
	\centering
	\includegraphics[scale= 0.4]{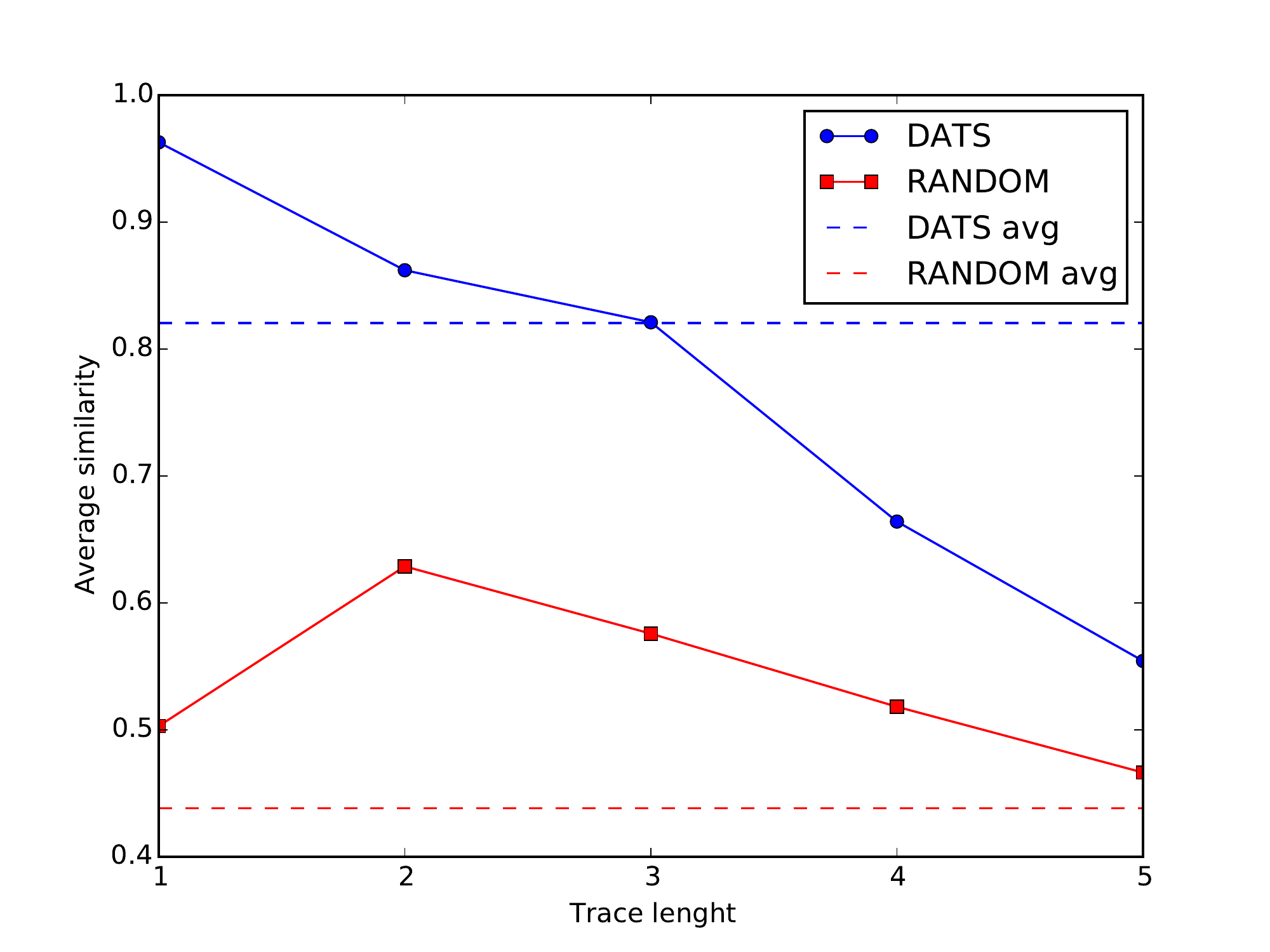}
	\caption{Activity sequence prediction results plot using the Damerau-Levenshtein similarity: the transition system is created using the sequence abstraction}
	\label{fig:path1dam}
\end{figure}

\begin{figure}[h!]
	\centering
	\includegraphics[scale= 0.4]{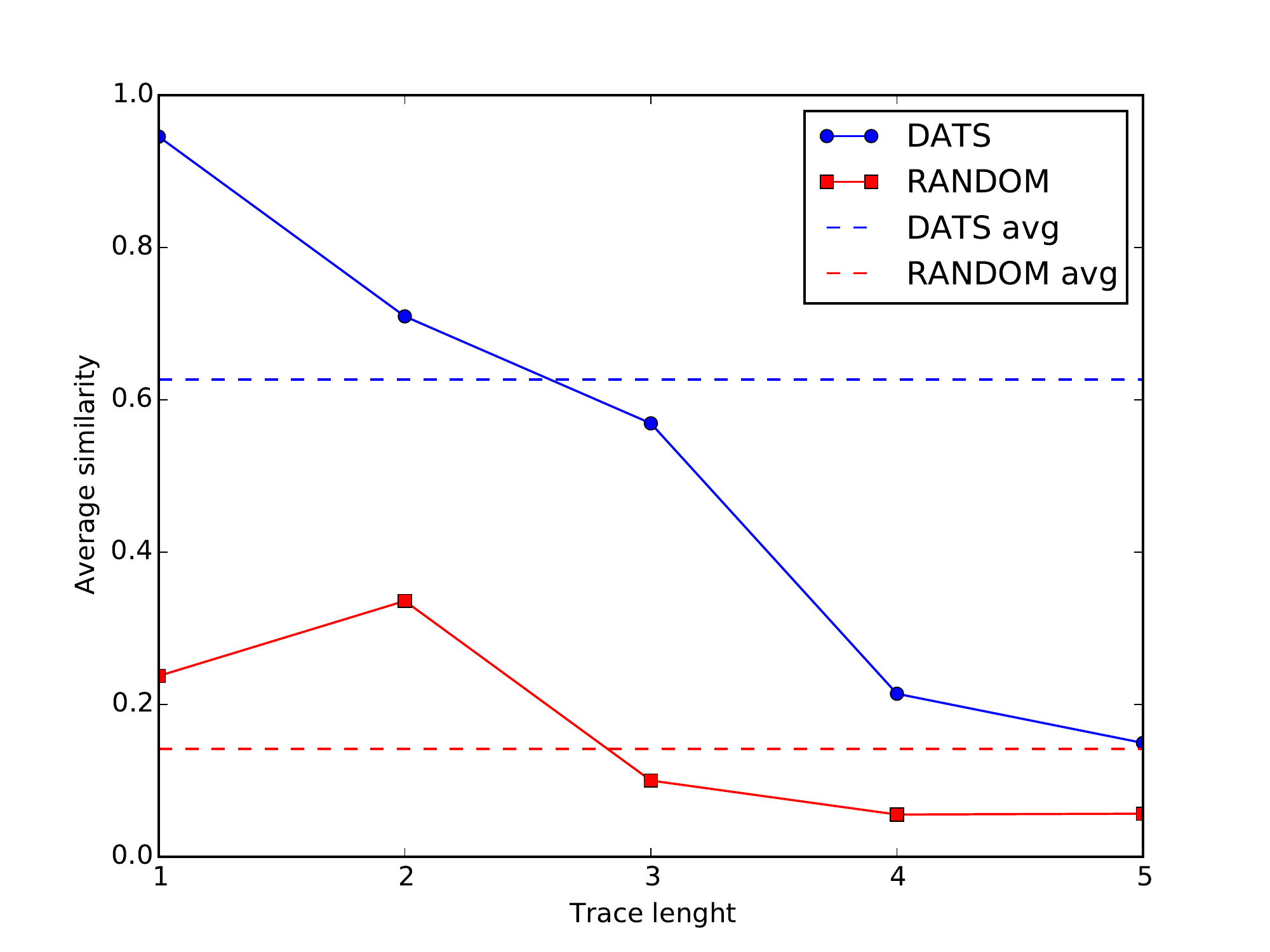}
	\caption{Activity sequence prediction results plot using the prefix similarity: the transition system is created using the sequence abstraction}
	\label{fig:path1pre}
\end{figure}

\begin{figure}[h!]
	\centering
	\includegraphics[scale= 0.4]{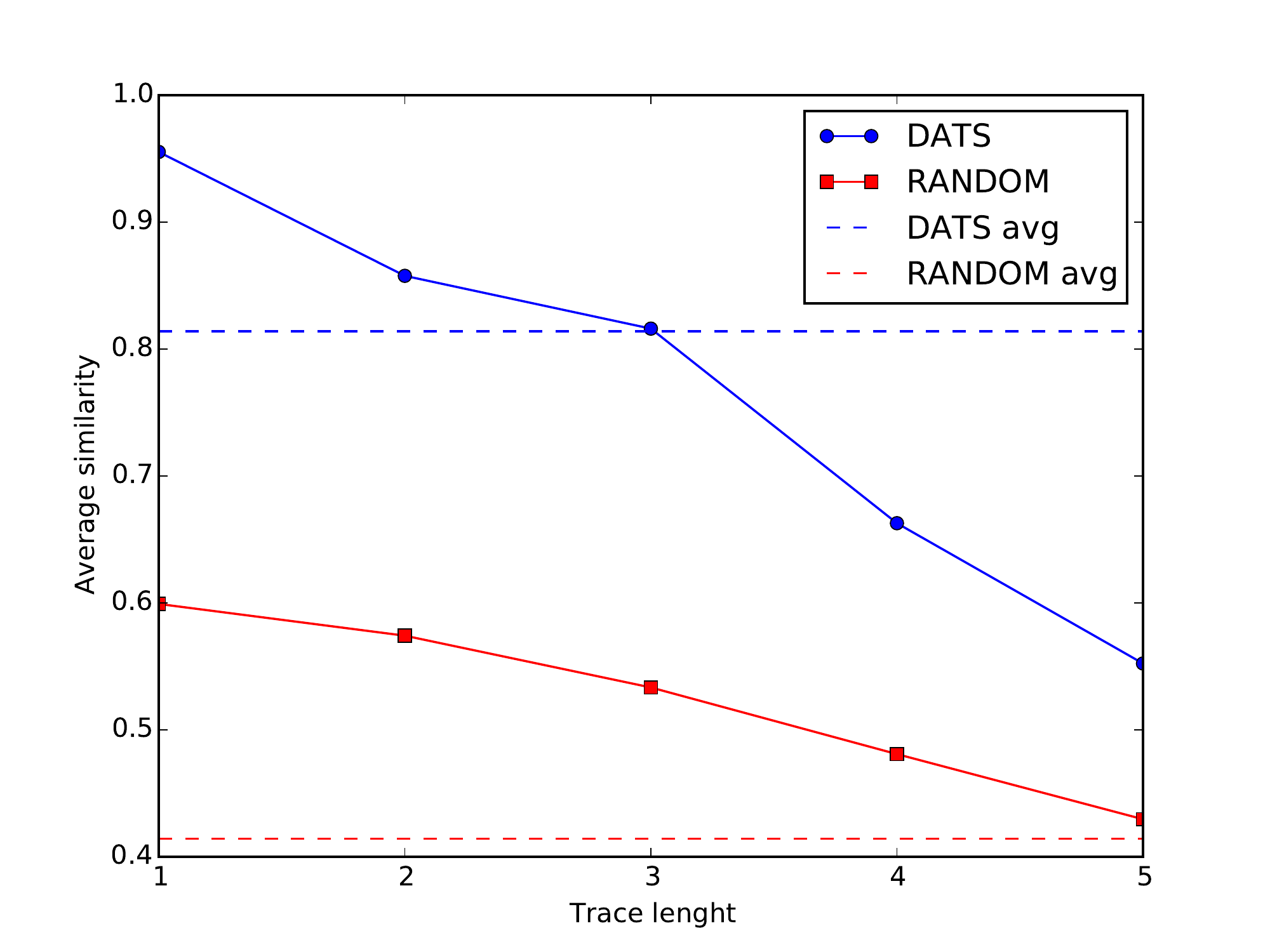}
	\caption{Activity sequence prediction results plot using the Damerau-Levenshtein similarity: the transition system is created using the multi-set abstraction}
	\label{fig:path2dam}
\end{figure}

\begin{figure}[h!]
	\centering
	\includegraphics[scale= 0.4]{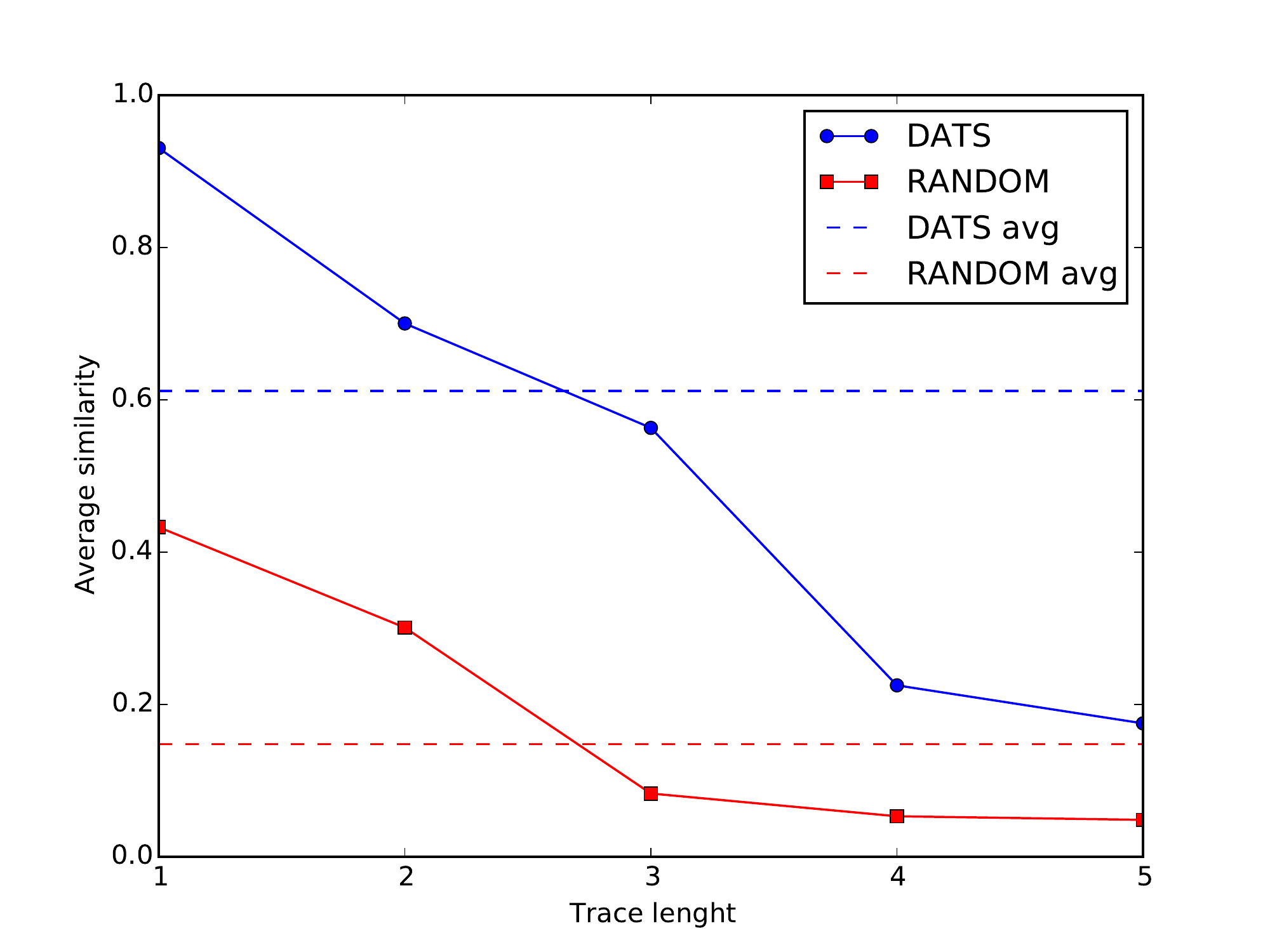}
	\caption{Activity sequence prediction results plot using the prefix similarity: the transition system is created using the multi-set abstraction}
	\label{fig:path2pre}
\end{figure}

\begin{figure}[h!]
	\centering
	\includegraphics[scale= 0.4]{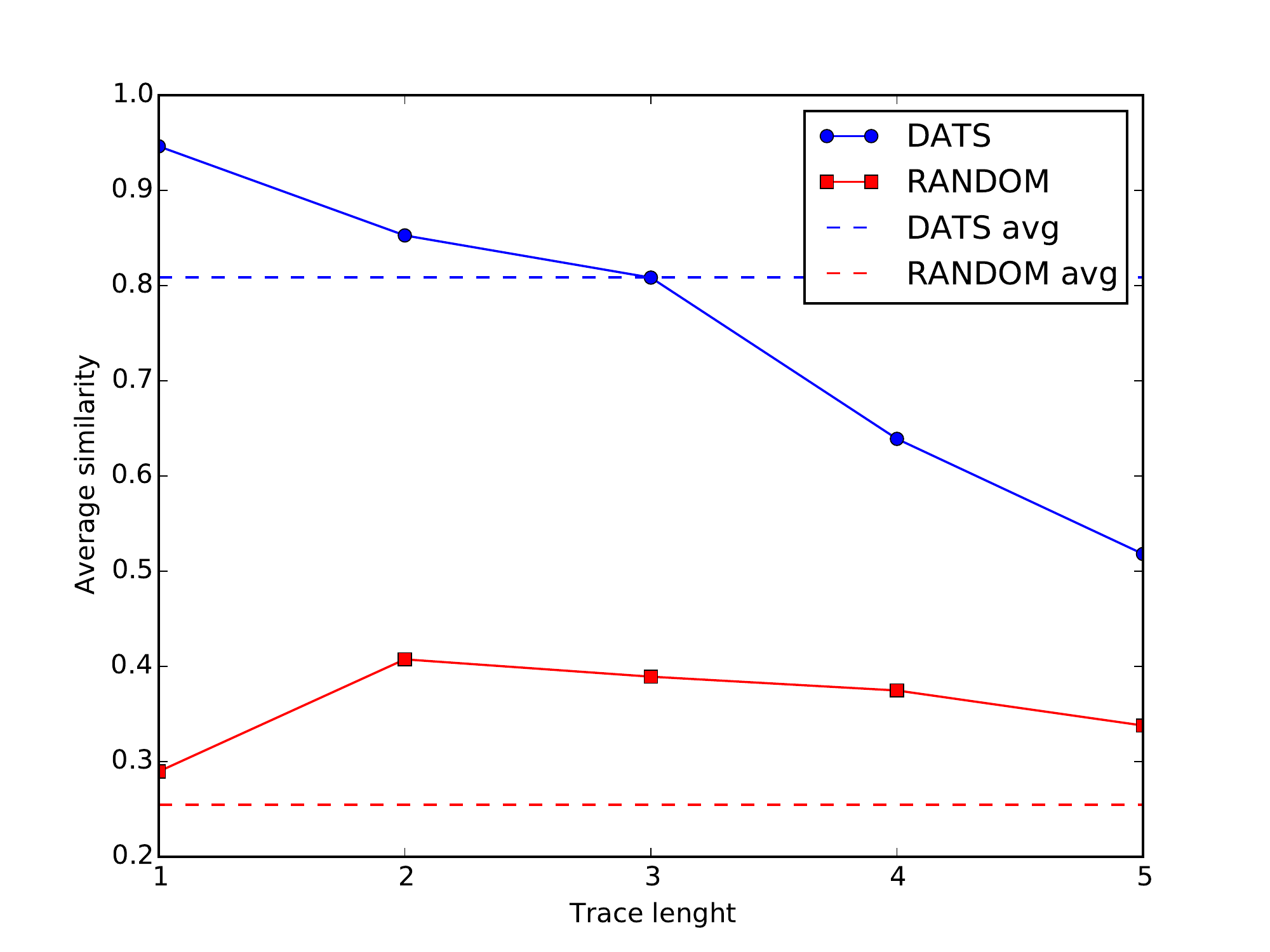}
	\caption{activity sequence prediction results plot using the Damerau-Levenshtein similarity: the transition system is created using the set abstraction}
	\label{fig:path3dam}
\end{figure}

\begin{figure}[h!]
	\centering
	\includegraphics[scale= 0.4]{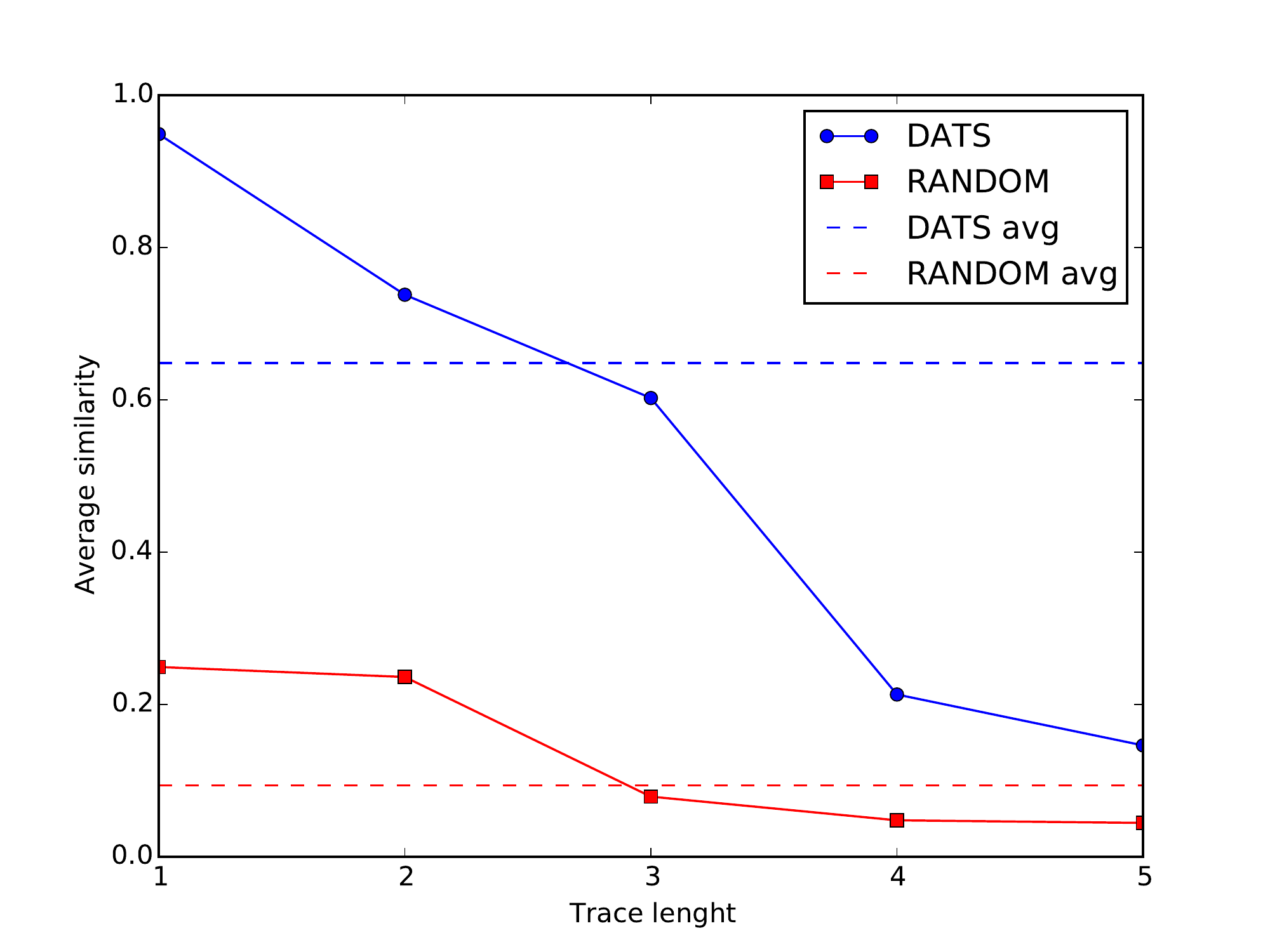}
	\caption{Activity sequence prediction results plot using the prefix similarity: the transition system is created using the set abstraction}
	\label{fig:path3pre}
\end{figure}

Using this event log we also test the future path prediction (FPP) method described in Section~\ref{sec:path}. A similar task is accomplished in \cite{Lakshmanan2013} using Markov chain, however they predict the likelihood of executing a specific activity in the future regardless the sequence of steps taken to get there. For this reason we decide to evaluates our method against a random predictor which chooses randomly the next activity according to the possible continuation seen in the event log. If an activity is also a possible termination the method randomly decides whether to stop or not.
As for the previous experiments, we used 5-fold cross validation. To evaluate the methods, which means assessing how much the predicted path respect the actual one, we used two metrics: the Damerau-Levenshtein similarity (DAM) and the common prefix (PRE). Tables~\ref{tab:path1}, ~\ref{tab:path2} and \ref{tab:path3} shows the results regarding the plots from Fig.~\ref{fig:path1dam} to Fig.~\ref{fig:path3pre}.

\begin{table}[h]
\centering
\begin{tabular}{l|cc|cc}
	\toprule
	&\multicolumn{2}{c|}{\textbf{FPP}}	&	\multicolumn{2}{c}{\textbf{RANDOM}} \\
	\#	&\textbf{DAM}	&\textbf{PRE}	&\textbf{DAM}	&\textbf{PRE}\\
	\midrule
	\textbf{1}	&0.9629$\pm$0.006	&0.9456$\pm$0.007	&0.5030$\pm$0.025	&0.2375$\pm$0.023	\\
	\textbf{2}	&0.8621$\pm$0.031	&0.7096$\pm$0.059	&0.6288$\pm$0.008	&0.3359$\pm$0.008	\\
	\textbf{3}	&0.8211$\pm$0.014	&0.5691$\pm$0.023	&0.5758$\pm$0.003	&0.1001$\pm$0.009	\\
	\textbf{4}	&0.6641$\pm$0.027	&0.2141$\pm$0.032	&0.5182$\pm$0.016	&0.0555$\pm$0.009	\\
	\textbf{5}	&0.5543$\pm$0.011	&0.1495$\pm$0.005	&0.4664$\pm$0.010	&0.0566$\pm$0.006	\\
	\textbf{$\mathbb{E}_\#$}	&0.8205$\pm$0.017	&0.6265$\pm$0.032	&0.4382$\pm$0.004	&0.1415$\pm$0.005	\\
	\bottomrule
\end{tabular}
\caption{Activity sequence prediction results: the transition system is created using the sequence abstraction.}
\label{tab:path1}
\end{table}

\begin{table}[h]
\centering
\begin{tabular}{l|cc|cc}
	\toprule
	&\multicolumn{2}{c|}{\textbf{FPP}}	&	\multicolumn{2}{c}{\textbf{RANDOM}} \\
	\#	&\textbf{DAM}	&\textbf{PRE}	&\textbf{DAM}	&\textbf{PRE}\\
	\midrule
	\textbf{1}& 0.9553$\pm$0.006 &0.9306$\pm$0.009 &0.5992$\pm$0.005 &0.4328$\pm$0.016 \\
	\textbf{2}&0.8577$\pm$0.039 &0.7003$\pm$0.071 &0.5742$\pm$0.011 &0.3010$\pm$0.012 \\
	\textbf{3}&0.8161$\pm$0.025 &0.5632$\pm$0.031 &0.5334$\pm$0.007 &0.0831$\pm$0.008 \\
	\textbf{4}&0.6628$\pm$0.029 &0.2251$\pm$0.038 &0.4809$\pm$0.012 &0.0532$\pm$0.005 \\
	\textbf{5}&0.5523$\pm$0.019 &0.1752$\pm$0.056 &0.4296$\pm$0.014 &0.0485$\pm$0.006 \\
	\textbf{$\mathbb{E}_\#$}&0.8140$\pm$0.027 &0.6118$\pm$0.045 &0.4143$\pm$0.003 &0.1477$\pm$0.003 \\
	\bottomrule
\end{tabular}
\caption{Activity sequence prediction results: the transition system is created using the multi-set abstraction.}
\label{tab:path2}
\end{table}

\begin{table}[h!]
\centering
\begin{tabular}{l|cc|cc}
	\toprule
	&\multicolumn{2}{c|}{\textbf{FPP}}	&	\multicolumn{2}{c}{\textbf{RANDOM}} \\
	\#	&\textbf{DAM}	&\textbf{PRE}	&\textbf{DAM}	&\textbf{PRE}\\
	\midrule
	\textbf{1}& 0.9463$\pm$0.012 &0.9489$\pm$0.005 &0.2896$\pm$0.017 &0.2492$\pm$0.024 \\
	\textbf{2}& 0.8527$\pm$0.007 &0.7380$\pm$0.012 &0.4074$\pm$0.006 &0.2361$\pm$0.012 \\
	\textbf{3}& 0.8084$\pm$0.008 &0.6024$\pm$0.014 &0.3892$\pm$0.006 &0.0790$\pm$0.006 \\
	\textbf{4}& 0.6390$\pm$0.004 &0.2131$\pm$0.011 &0.3747$\pm$0.007 &0.0479$\pm$0.004 \\
	\textbf{5}& 0.5181$\pm$0.008 &0.1463$\pm$0.006 &0.3379$\pm$0.015 &0.0445$\pm$0.010 \\
	\textbf{$\mathbb{E}_\#$}& 0.8088$\pm$0.005 &0.6484$\pm$0.004 &0.2548$\pm$0.001 &0.0936$\pm$0.002 \\
	\bottomrule
\end{tabular}
\caption{Activity sequence prediction results: the transition system is created using the set abstraction.}
\label{tab:path3}
\end{table}

Each row in the tables represents the similarity considering sequences of exactly $n$ activities where $n$ is indicated in the $\#$ column. Rows with the $\mathbb{E}_\#$ symbol are the average similarities considering every sequence length.
Readers can notice that with every abstraction the proposed method outperforms the random one. In particular, FPP it is able to identify the next activity almost 94\% of the times and the similarity of the next two is almost 0.86 in average with an hit rate of 71\% in average. We achieve good results, with respect to Damerau-Levenshtein similarity, even with 3, 4 and 5 activity sequences. 

The trend of the plots look similar and we can notice how the difference in the prediction accuracy are evident starting on the left side (few activities in the future) and narrow on the right side. This fact is due to the stockpile of the uncertainty of each step in the future, which makes the prediction of the FPP closer to the random one.

However, in average, our approach obtained for DAM and PRE 0.81 and 0.63 respectively, which are far better than the 0.36 and 0.09 obtained by the random method. 

\section{Conclusions}
\label{sec:conc}

In this work we presented some new methods that can be employed to tackle the problem of predicting the time-to-completion of running business process instances. The contributions we presented in this work can be summarized as the following:
\begin{itemize}
	\item we proposed three new prediction methods, which take advantage not only of the control-flow information but also of the additional data presented in the event log;
	\item we leveraged on solid and well-studied machine learning techniques in order to build models able to manage the additional information;
	\item we constructed our approach in order to deal with unexpected behaviours or noisy data, by looking at the closeness between the new trace and the most similar process flows already observed;
	\item we have extensively evaluated our algorithms on real life data and show that our methods outperform state-of-the-art.
\end{itemize} 

The proposed set of methods aims to face different scenarios and to overcome the limitations of the state-of-the-art approaches. Moreover, we distinguished the application of the prediction problem into two main scenarios:
\begin{enumerate}
	\item the process is stable and consequently the event log used to train the model contains all the possible process behaviours;
	\item the process is dynamic (i.e., might contain drifts) and, consequently, the event log used for training does not contain all the possible process behaviours, e.g., seasonability of the process.
\end{enumerate}
Experiments in \cite{Polato2014} and those reported in Section~\ref{sec:implementation} have shown how the DATS approach overcome the state-of-the-art in the first scenario. Assuming that the training event log contains all the possible process' behaviours, it is possible to take advantage from the static nature of the process and rely on the transition system structure. A central role in this method is played by the Na\"ive Bayes classifiers, which are also involved in the prediction of future sequence of activities in Section~\ref{sec:path}. These two tools together can be very useful for business managers because, in addition to the remaining time estimation, they have also some hints about the sequence of activities that the instance is going to take. Using these information business managers can act preventively and they can try to avoid uncomfortable situations.
However, we also obtained good result with the single SVR-based methods which are well suited for the second scenario in which not all the process' behaviours are present in the training phase. Here, the SVR+TS method is not affected by the lack of information in the training set because is able to generalize the workflow thanks to the similarity-based transition system and the nature itself of the single-SVR approach. Experimental results point out that methods which are strongly dependent on the TS structure have problem with new process' variants, while SVR method with the similarity-based TS, outperforms all the other approaches.

\vspace{1em}
As future work, we plan to improve the parameters calibration of our approaches, in order to improve the overall results of our approach. We would like to investigate whether taking into account only work-hours in the prediction is valuable or not. Moreover, we would like to deploy our approach on real scenario, in order to stress the whole approach under production-level constraints.

\section*{Acknowledgment}
The work reported in this paper is supported by the Eurostars-Eureka project PROMPT (E!6696).



\bibliographystyle{elsarticle-num}
\bibliography{library,library2}

\end{document}